%% file: root_combined.tex
\documentclass[letterpaper,10pt,journal,twoside]{ieeetran}
\input{header}

\makeatletter
\let\NAT@parse\undefined
\makeatother
\usepackage[colorlinks]{hyperref}
\usepackage[capitalize]{cleveref}

\usepackage{multibib}
\newcites{app}{References}  %

\newcommand{\shorttitle}{Learning from Guided Play}

\ifhighlight
    \title{Learning from Guided Play:\\ Improving Exploration for Adversarial Imitation Learning with Simple Auxiliary Tasks \hl{(revisions)}}
\else
    \title{Learning from Guided Play:\\ Improving Exploration for Adversarial Imitation Learning with Simple Auxiliary Tasks}
\fi

\author{Trevor Ablett$^{1}$, Bryan Chan$^{2}$, and Jonathan Kelly$^{1}$
\thanks{Manuscript received: Nov.\ 3, 2022; Accepted: Dec.\ 18, 2022.}  %
\thanks{This paper was recommended for publication by Editor Jens Kober upon evaluation of the Associate Editor and Reviewers' comments.}  %
\thanks{$^{1}$Authors are with the Space \& Terrestrial Autonomous Robotic Systems (STARS) Laboratory at the University of Toronto Institute for Aerospace Studies (UTIAS), Toronto, Ontario, Canada, M3H~5T6. Email: {\tt\small <first name>.<last name>@robotics.utias.utoronto.ca}}
\thanks{$^{2}$Author is with the Department of Computing Science at the University of Alberta, Edmonton, Alberta, Canada, T6G 2E8. Email: {\tt\small bryan.chan@ualberta.ca}}
\thanks{Digital Object Identifier (DOI): see top of this page.}  %
}

\begin{document}

\maketitle

\markboth{IEEE ROBOTICS AND AUTOMATION LETTERS. PREPRINT VERSION. ACCEPTED DEC, 2022}
{Ablett \MakeLowercase{\textit{et al.}}: \shorttitle}

\input{sections/0-abstract}

\begin{IEEEkeywords}
Imitation Learning, Reinforcement Learning, Transfer Learning
\end{IEEEkeywords}

\input{sections/1-introduction}
\input{sections/3-methodology}

\input{sections/4-experiments}
\input{sections/5-analysis}

\input{sections/2-background}

\input{sections/6-conclusion}

\input{sections/acknowledgements}

\bibliographystyle{IEEEtran}
\bibliography{lfgp}

\newpage
\input{sections/appendix}

\bibliographystyleapp{IEEEtran}
\bibliographyapp{lfgp}

\end{document}

%% file: header.tex
\IEEEoverridecommandlockouts
\pdfoutput=1

\input{packages}

\input{math_shortcuts}

%% file: packages.tex
\usepackage{tabularx}
\usepackage{booktabs}
\usepackage{multirow}
\usepackage{amsmath}
\usepackage{amssymb}
\usepackage[noadjust]{cite}
\usepackage{xparse} 
\usepackage{amsfonts} 
\usepackage{bbm}
\usepackage{color}
\usepackage{mathtools}
\usepackage{graphicx}
\usepackage{subcaption}
\usepackage{calc}
\usepackage{balance}
\usepackage{url}
\usepackage{algorithm}
\usepackage{algorithmic}

\usepackage{xcolor}
\newcommand\hl[1]{%
  \bgroup
  \hskip0pt\color{green!80!black}%
  #1%
  \egroup
}

\newif\ifhighlight

\ifhighlight
\else
    \renewcommand\hl[1]{#1}
\fi

%% file: math_shortcuts.tex
\usepackage{xparse}

\newcommand{\mdpm}{\mathcal{M}}
\newcommand{\mdps}{\mathcal{S}}
\newcommand{\mdpa}{\mathcal{A}}

\newcommand{\mdpp}{\mathcal{P}}
\newcommand{\mdprho}{\rho_0}

\newcommand{\buffer}{\mathcal{B}}
\newcommand{\tasks}{\mathcal{T}}

\usepackage{color}

\NewDocumentCommand\bbm{}{ \begin{bmatrix} }
\NewDocumentCommand\ebm{}{ \end{bmatrix} }

\NewDocumentCommand\Real{}{ \mathbb{R} }

%% file: sections/0-abstract.tex
\begin{abstract}
Adversarial imitation learning (AIL) has become a popular alternative to supervised imitation learning that reduces the distribution shift suffered by the latter.
However, AIL requires effective exploration during an online reinforcement learning phase.
In this work, we show that the standard, na\"ive approach to exploration can manifest as a suboptimal local maximum if a policy learned with AIL sufficiently matches the expert distribution without fully learning the desired task.
This can be particularly catastrophic for manipulation tasks, where the difference between an expert and a non-expert state-action pair is often subtle.
We present Learning from Guided Play (LfGP), a framework in which we leverage expert demonstrations of multiple exploratory, auxiliary tasks in addition to a main task.
The addition of these auxiliary tasks forces the agent to explore states and actions that standard AIL may learn to ignore.
Additionally, this particular formulation allows for the reusability of expert data between main tasks. %
Our experimental results in a challenging multitask robotic manipulation domain indicate that LfGP significantly outperforms both AIL and behaviour cloning, while also being more expert sample efficient than these baselines. 
To explain this performance gap, we provide further analysis of a toy problem that highlights the coupling between a local maximum and poor exploration, and also visualize the differences between the learned models from AIL and LfGP.\footnote[3]{Code, Blog, Appendix: \url{https://papers.starslab.ca/lfgp}\label{fn:supp}}
\end{abstract}

%% file: sections/1-introduction.tex
\section{Introduction} \label{sec:intro}
\IEEEPARstart{E}{xploration} is a crucial part of effective reinforcement learning (RL). %
A variety of methods have attempted to optimize the exploration-exploitation trade-off of RL agents \cite{suttonReinforcementLearningIntroduction2018, bellemareUnifyingCountBasedExploration2016, nairOvercomingExplorationReinforcement2018}, but the development of a technique that generalizes across domains remains an open research problem. %
A simple, well-known approach to reduce the need for random exploration is to provide a dense, or ``shaped,'' reward to learn from, but this can be very challenging to design appropriately \cite{ngShapingPolicySearch2003}.
Furthermore, the environment may not directly provide the low-level state information required for such a reward.
An alternative to providing a dense reward is to learn a reward function from expert demonstrations of a task, in a process known as inverse RL (IRL) \cite{ngAlgorithmsInverseReinforcement2000}.
Many modern approaches to IRL are part of the adversarial imitation learning (AIL) family \cite{hoGenerativeAdversarialImitation2016}.
In AIL, rather than learning a reward function directly, the policy and a learned discriminator form a two-player min-max optimization problem, where the policy aims to confuse the discriminator by producing expert-like data, while the discriminator attempts to classify expert and non-expert data.

\begin{figure}[t]
	\centering
	\includegraphics[width=\linewidth]{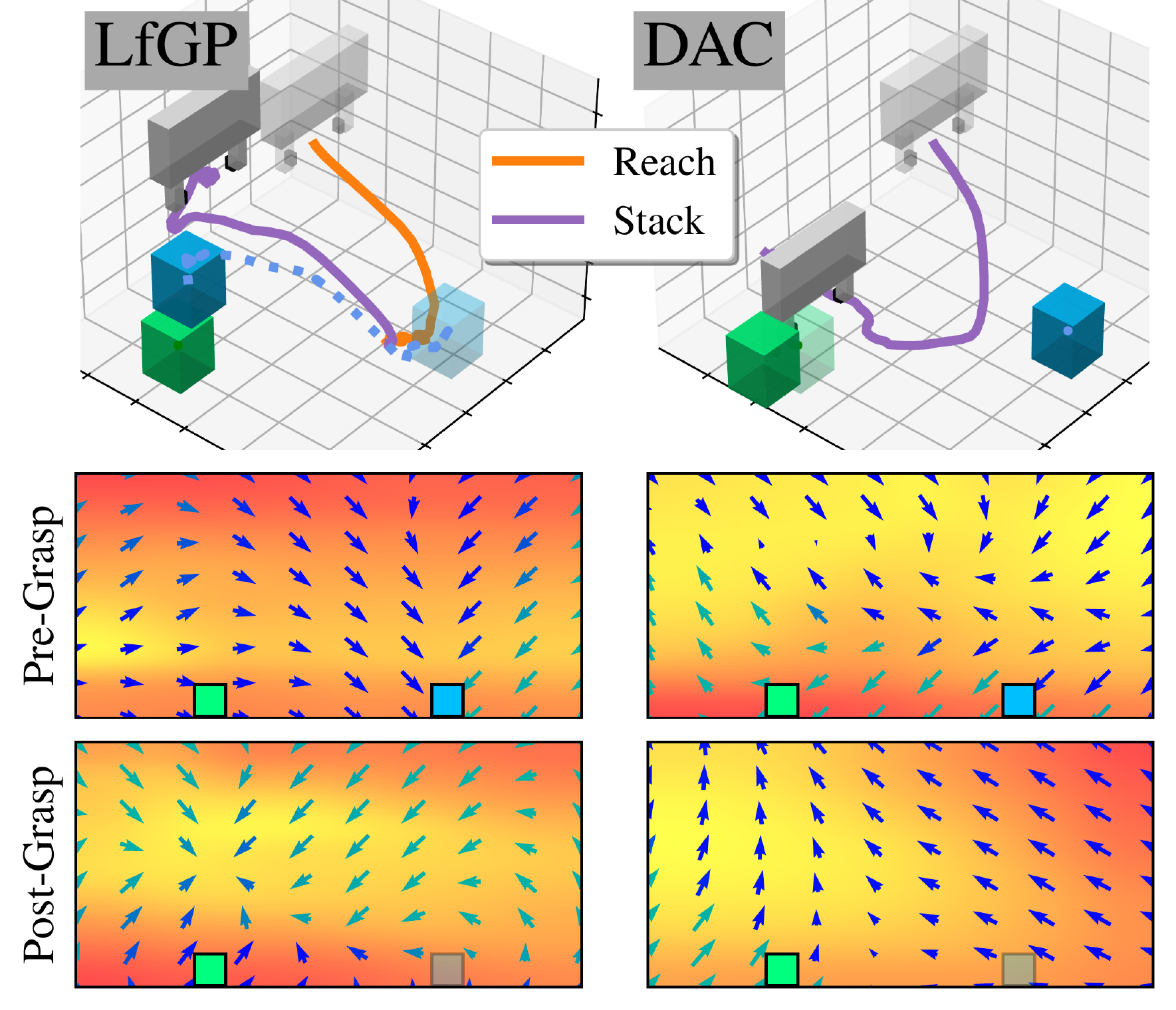}
	\caption{Learning from Guided Play (LfGP) finds an effective stacking policy by learning to compose multiple simple auxiliary tasks (only Reach is shown, for this episode) along with stacking.
	Discriminator Actor-Critic (DAC) \cite{kostrikovDiscriminatorActorCriticAddressingSample2019}, or off-policy AIL, reaches a local maximum action-value function and policy, failing to solve the task.
	Arrow direction indicates mean policy velocity action, red-to-yellow (background) indicates low-to-high learned value, while arrow colour indicates probability of closing (green) or opening (blue) the gripper. }
	\label{fig:lfgp_vs_dac_traj_and_q}
	\vspace{-6mm}
\end{figure}

\begin{figure*}[ht]
    \vspace{2mm}
	\centering
	\includegraphics[width=.85\linewidth]{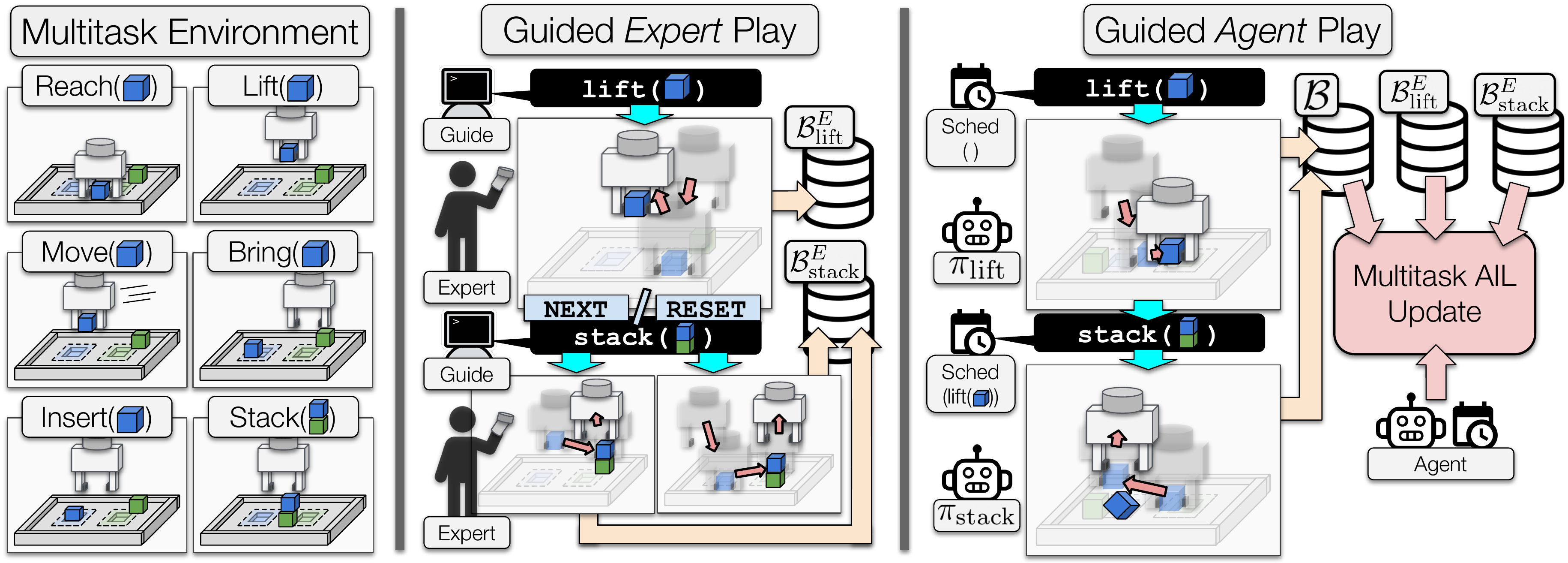}
	\caption{The main components of our system for learning from guided play. In a multitask environment, a guide prompts an expert for a mix of multitask demonstrations, after which we learn a multitask policy through scheduled hierarchical AIL.}
	\label{fig:system}
	\vspace{-4mm}
\end{figure*}

Although AIL has been shown to be more \textit{expert} sample efficient than supervised imitation learning (also known as behavioural cloning, or BC) in continuous-control environments \cite{hoGenerativeAdversarialImitation2016,fuLearningRobustRewards2018,kostrikovDiscriminatorActorCriticAddressingSample2019}, its application to long-horizon robotic manipulation tasks with a wide distribution of possible initial configurations remains challenging \cite{kostrikovDiscriminatorActorCriticAddressingSample2019, orsiniWhatMattersAdversarial2021}.
In this work, we investigate the use of AIL in a multitask robotic manipulation domain. %
We find that a state-of-the-art AIL method, in which off-policy learning is used to maximize \textit{environment} sample efficiency \cite{kostrikovDiscriminatorActorCriticAddressingSample2019} (i.e., reduce the quantity of environment interaction required from the online RL portion of AIL), is outperformed by BC with an equivalent amount of expert data, contradicting previous results \cite{hoGenerativeAdversarialImitation2016,fuLearningRobustRewards2018,kostrikovDiscriminatorActorCriticAddressingSample2019}.
Through a simplified example, simulated robotic experiments, and learned model analysis, we show that this outcome occurs because a model learned with expert data and a discriminator is susceptible to the deceptive reward problem \cite{ecoffetFirstReturnThen2021}.
In other words, while AIL, and more generally IRL, can provide something akin to a dense reward, this reward is not necessarily optimal for teaching, and AIL alone does not enforce sufficiently diverse exploration to escape locally optimal but globally poor models.
A locally-optimal policy has converged to match a subset of the expert data, but in doing so, avoids crucial states and actions (e.g., in \cref{fig:lfgp_vs_dac_traj_and_q}, grasping the blue block) required to globally match the full expert set.

To overcome this limitation of AIL, we present Learning from Guided Play (LfGP),\footnote[4]{Originally presented as a non-archival workshop paper \cite{ablettLearningGuidedPlay2021}.} in which we combine AIL with a scheduled approach to hierarchical RL (HRL) \cite{riedmillerLearningPlayingSolving2018}, allowing an agent to `play' in the environment with an expert guide.
Using expert demonstrations of multiple relevant auxiliary tasks (e.g., Reach, Lift, Move-Object), along with a main task (e.g., Stack, Bring, Insert), our scheduled hierarchical agent is able to learn tasks where AIL alone fails.
Crucially, our formulation also allows auxiliary expert data to be reused between main tasks, further emphasizing the expert sample efficiency of our method.

We use the word \textit{play} to describe an agent that simultaneously attempts and learns numerous tasks at once, freely composing them together, inspired by the playful (as opposed to goal-directed) phase of learning experienced by children \cite{riedmillerLearningPlayingSolving2018}.
In our case, \textit{guided} represents two separate but related ideas: first, that the expert guides this play, as opposed to requiring hand-crafted sparse rewards as in \cite{riedmillerLearningPlayingSolving2018} (right side of \cref{fig:system}), and second, that the expert gathering of multitask, semi-structured demonstrations is \textit{guided} by uniform-random task selection (middle of \cref{fig:system}), rather than requiring the expert to choose transitions between goals, as in \cite{lynchLearningLatentPlans2019, guptaRelayPolicyLearning2019}.
Our specific contributions are the following:
\begin{enumerate}
    \item A novel application of a hierarchical framework \cite{riedmillerLearningPlayingSolving2018} to AIL that learns a reward and policy for a challenging main task by simultaneously learning rewards and policies for auxiliary tasks.
    \item Manipulation experiments in which we demonstrate that AIL fails, while LfGP significantly outperforms both AIL and BC.
    \item A thorough ablation study to examine the effects of various design choices for LfGP and our baselines.
    \item Empirical analysis, including a simplified representative example and visualization of the learned models of LfGP and AIL, to better understand why AIL fails and how LfGP improves upon it.
\end{enumerate}

%% file: sections/3-methodology.tex
\section{Problem Formulation}

A Markov decision process (MDP) is defined as $\mdpm = \langle \mdps, \mdpa, R, \mdpp, \mdprho, \gamma \rangle$, where the sets $\mdps$ and $\mdpa$ are respectively the state and action space, $R : S \times A \rightarrow \Real$ is a reward function, $\mdpp$ is the state-transition environment dynamics distribution, $\mdprho$ is the initial state distribution, and $\gamma$ is the discount factor. 
Actions are sampled from a stochastic policy $\pi(a|s)$.
The policy $\pi$ interacts with the environment to yield experience $\left( s_t, a_t, r_t, s_{t + 1} \right)$ for $t = 0, \dots, \infty$,  where $s_0 \sim \mdprho(\cdot), a_t \sim \pi(\cdot | s_t), s_{t+1} \sim \mdpp(\cdot | s_t, a_t), r_t = R(s_t, a_t)$.
When referring to finite-horizon tasks, $t=T$ indicates the final timestep of a trajectory.

For notational convenience, we assume infinite-horizon, non-terminating environments where $t$ is unbounded, but the extension to the finite-horizon case is trivial.
We aim to learn a policy $\pi$ that maximizes the expected return
$J(\pi) = \mathbb{E}_{\pi}\left[ G(\tau_{0:\infty}) \right] = \mathbb{E}_\pi\left[ \sum_{t=0}^{\infty} \gamma^{t} R(s_t, a_t) \right]$,
where $\tau_{t:\infty} = \{(s_t, a_t), \dots\}$ is the trajectory starting with $(s_t, a_t)$, and $G(\tau_{t:\infty})$ is the return of trajectory $\tau$. 

In this work, we focus on imitation learning (IL), where $R$ is unknown and instead we are given a finite set of expert demonstration $(s,a)$ pairs $\buffer^E = \left\{ (s,a)^{E}, \dots \right\}$.
In AIL, we attempt to simultaneously learn $\pi$ and a discriminator $D : S \times A \rightarrow \left[0, 1 \right]$ that differentiates between expert samples $(s,a)^E$ and policy samples $(s,a)^\pi$ and subsequently define $R$ using $D$ \cite{hoGenerativeAdversarialImitation2016, kostrikovDiscriminatorActorCriticAddressingSample2019}.
To accommodate hierarchical learning, we augment $\mdpm$ to contain auxiliary tasks, where $\tasks_{\text{aux}} = \left\{ \tasks_1, \dots, \tasks_K \right\}$ are separate MDPs that share $\mdps, \mdpa, \mdpp, \mdprho$ and $\gamma$ with the main task $\tasks_{\text{main}}$ but have their own reward functions, $R_{k}$. 
With this modification, we refer to entities in our model that are specific to task $\tasks \in \tasks_{\text{all}}$, $\tasks_\text{all} = \tasks_\text{aux} \cup \left\{ \tasks_\text{main} \right\}$, as $(\cdot)_\tasks$.
We assume that we have a set of expert data $\buffer^E_\tasks$ for each task.

\section{Local Maximum with Off-Policy AIL}
\label{sec:local_maximum_ail}

\begin{figure}[t]
    \centering
    \includegraphics[width=.9\linewidth]{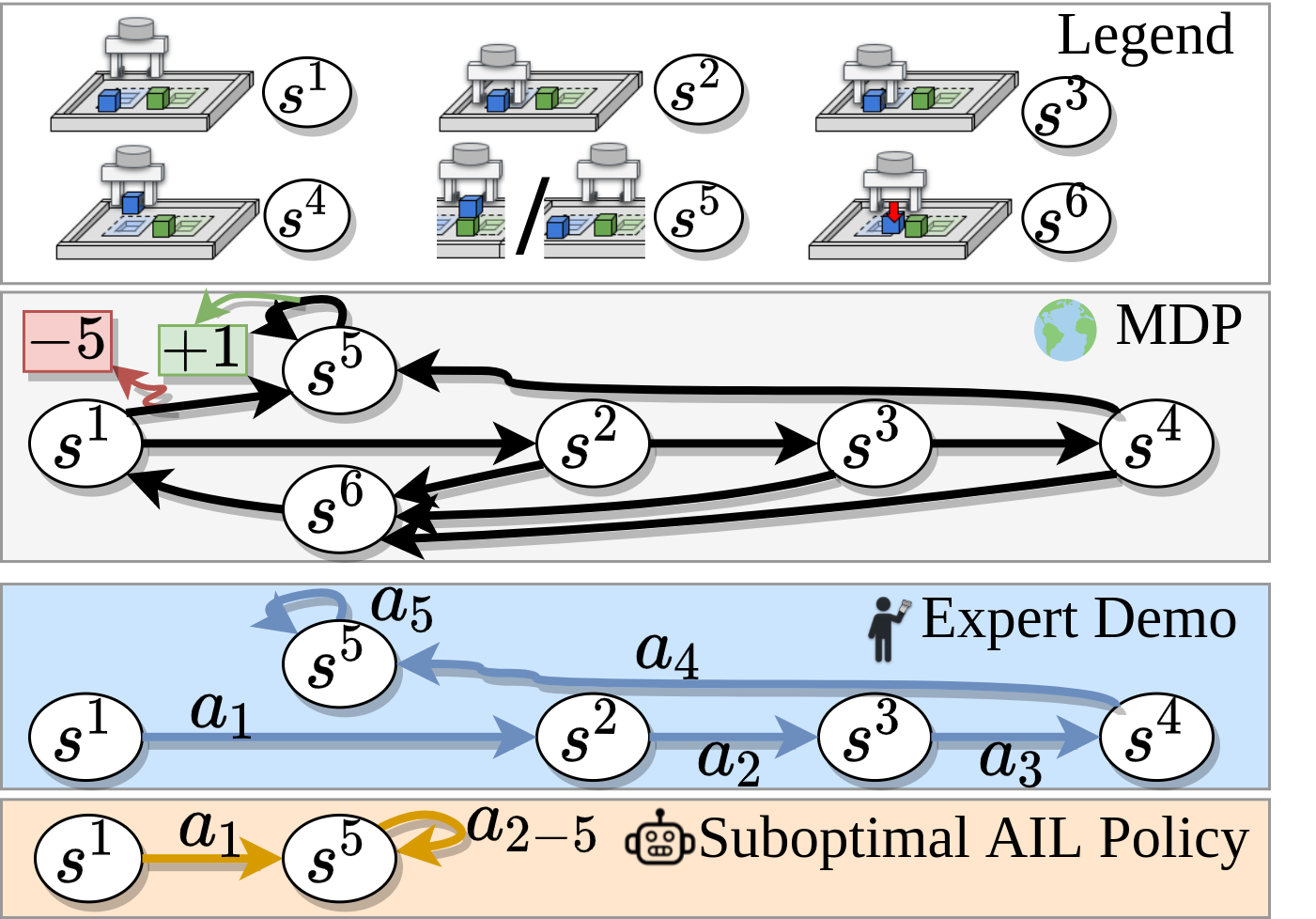}
    \caption{An MDP, analogous to stacking, with an expert demonstration. Poor exploration can lead AIL to learn a suboptimal policy.}
    \label{fig:six_state_toy}
    \vspace{-4mm}
\end{figure}

In this section, we provide a representative example of how AIL can fail by reaching a locally maximum policy due to a learned deceptive reward \cite{ecoffetFirstReturnThen2021} coupled with poor exploration.
A simple six-state MDP is shown in \cref{fig:six_state_toy}, with ten state-conditional actions.
We refer to actions as $a_t = a^{nm}$ and states as $s_t = s^{n}$ where $t$, $n$ and $m$ refer to the current timestep, current state, and next state, respectively.
The reward function is $R(s^5, a^{55}) = +1$, $R(s^1, a^{15}) = -5$ and 0 for all other state-action pairs.
The initial state $s_1$ is always $s^1$, the fixed horizon length is 5, and no discounting is used.

The MDP is meant to be roughly analogous to a stacking manipulation task: $s^2, s^3, s^4$ and $s^6$ represent the first block being reached, grasped, lifted, and dropped respectively. 
State $s^5$ represents the gripper hovering over the second block (whether the first block has been stacked or not), while $s^1$ is the reset state, and $a^{15}$ represents reaching $s^5$ without grasping the first block.
Taking action $a^{15}$ results in a total return of -1 (because $R(s^1, a^{15}) = -5$), since the first block has not actually been grasped.
In our case, the agent does not receive any reward, and instead an expert demonstration of the optimal trajectory is provided.
We will assume access to a learned (perfect) discriminator, and will use the AIRL \cite{fuLearningRobustRewards2018} reward, so state-action pairs in the expert set receive +1 reward and all others receive -1.

We define the action-value $Q(s_t, a_t)$ as the expected value of taking action $a_t$ in state $s_t$, and initialize it to zero for all $(s,a)$ pairs.
We define our update rule as the standard Q-Learning update \cite{suttonReinforcementLearningIntroduction2018}, $Q(s_t, a_t) = Q(s_t, a_t) + \alpha \left( R(s_t, a_t) + \max_a Q(s_{t+1}, a) - Q(s_t, a_t) \right)$, with $\alpha = 0.1$.
The agent uses $\epsilon$-greedy exploration, storing each $(s_t, a_t, s_{t+1})$ tuple into a buffer.
After each episode, all Q values are updated to convergence using the whole buffer.

After the first complete episode of $\{a^{15}, a^{55}, a^{55}, a^{55}, a^{55} \}$, $Q(s^1, a^{15}) = 2.7$, and $Q(s^1, a^{12}) = 0$.
In the second ($\{a^{12}, a^{26}, a^{61}, a^{15}, a^{55} \}$) and third ($\{a^{12}, a^{23}, a^{36}, a^{61}, a^{15} \}$) episodes, the agent initially moves in the correct direction, but ultimately still fails. 
The final Q values in $s^1$ are $Q(s^1, a^{15}) = 0.49$ and $Q(s^1, a^{12}) = 0.13$.\footnote[5]{\hl{See \texttt{six\_state\_mdp.py} from open source code to reproduce.}}

A policy maximizing Q, having simultaneously learned to avoid $s^6$ (by avoiding $s^2$ and $s^3$) and exploiting the $(s^5, a^{55})$ expert pair, will choose $a_1 = a^{15}$, giving a final return of\linebreak -1 in the real MDP.
This behaviour matches what we see in \cref{fig:lfgp_vs_dac_traj_and_q}: due to the large negative reward from dropping the block, AIL learns a policy that avoids stacking altogether and merely reaches the second block, just as AIL here learns to skip $s^2$ and $s^3$ and exploit $a^{55}$.
In both cases, poor initial exploration leads to a deceptive reward, which exacerbates  poor exploration.

\section{Learning from Guided Play (LfGP)}
\label{sec:method}

We now introduce Learning from Guided Play (LfGP). %
Our primary goal is to learn a policy $\pi_{\tasks_{\text{main}}}$ that can solve the main task $\tasks_{\text{main}}$, with a secondary goal of also learning auxiliary task policies $\pi_{\tasks_{\text{1}}}, \dots, \pi_{\tasks_{\text{K}}}$ that are used for improved exploration.
More specifically, we derive a hierarchical learning objective that is decomposed into three parts: i) recovering the reward function of each task with expert demonstrations, ii) training all policies to achieve their respective goals, and iii) using all policies for effective exploration in $\tasks_\text{main}$.
For a summary of the algorithm, see supplementary material link in \cref{fn:supp}.

\subsection{Learning the Reward Function}
We first describe how to recover the reward functions from expert demonstrations. 
For each task $\tasks \in \tasks_\text{all}$, we learn a discriminator $D_{\tasks}(s, a)$ that is used to define the reward function for policy optimization. 
We construct the joint discriminator loss following \cite{kostrikovDiscriminatorActorCriticAddressingSample2019} to train each discriminator in an off-policy manner:

\begin{equation}
\label{DAC}
\begin{aligned}
    \mathcal{L}(D) = -\sum_{\tasks \in \tasks_\text{all}} \mathbb{E}_{\buffer}\left[ \log \left( 1 - D_\tasks(s, a)\right) \right]\\ + \mathbb{E}_{\buffer^E_{\tasks}}\left[ \log \left( D_\tasks(s, a)\right) \right].
\end{aligned}
\end{equation}
Each resulting discriminator $D_\tasks$ attempts to differentiate the occupancy measure between the distributions induced by $\buffer^E_\tasks$ and $\buffer$. 
We can use $D_\tasks$ to define various reward functions \cite{kostrikovDiscriminatorActorCriticAddressingSample2019}; following \cite{fuLearningRobustRewards2018}, we define the reward function for each task $\tasks$ to be $R_\tasks(s_t, a_t) = \log \left( D_\tasks(s_t, a_t) \right) - \log \left( 1 - D_\tasks(s_t, a_t) \right)$.

\subsection{Learning the Hierarchical Agent}
We adapt Scheduled Auxiliary Control (SAC-X) \cite{riedmillerLearningPlayingSolving2018} to learn the hierarchical agent. 
The agent includes low-level intention policies (equivalently referred to as intentions), a high-level scheduler policy, as well as the Q-functions and the discriminators.
The intentions aim to solve their corresponding tasks (i.e., the intention $\pi_\tasks$ aims to maximize the task return $J(\pi_\tasks)$), whereas the scheduler aims to maximize the expected return for $\tasks_\text{main}$ by selecting a sequence of intentions to interact with the environment.
For the remainder of the paper, when we refer to a policy, we are referring to an intention policy, as opposed to the scheduler, unless otherwise specified.

\subsubsection{Learning the Intentions}
We learn each intention using Soft Actor-Critic (SAC) \cite{haarnojaSoftActorCriticOffPolicy2018}, an actor-critic algorithm that maximizes the entropy-regularized objective, though any off-policy RL algorithm would suffice.
The objective is 
\begin{eqnarray}
    J(\pi_\tasks) = 
    \mathbb{E}_{\pi_\tasks}\left[ \sum_{t = 0}^{\infty} \gamma^{t} \left(R_\tasks(s_t, a_t) + \alpha\mathcal{H}(\pi_\tasks(\cdot | s_t))\right) \right],
\end{eqnarray}
where the learned temperature $\alpha$ determines the importance of the entropy term and $\mathcal{H}(\pi_\tasks (\cdot | s_t))$ is the entropy of the intention $\pi_\tasks$ at state $s_t$. 
The soft Q-function is 
\begin{equation}
\begin{aligned}
    Q_\tasks&(s_t, a_t) = R_\tasks(s_t, a_t) \\
    & + \mathbb{E}_{\pi_\tasks}\left[ \sum_{t = 0}^{\infty} \gamma^t (R_\tasks(s_{t+1}, a_{t+1}) + \alpha\mathcal{H}(\pi_\tasks(\cdot | s_{t+1}))) \right].
\end{aligned}
\label{q_func}
\end{equation}
The intentions maximize the joint policy objective 
\begin{equation}
    \label{complete_PI}
    \mathcal{L}(\pi_\text{int}) = \sum_{\tasks \in \tasks_\text{all}}\mathbb{E}_{s \sim \buffer_\text{all}, a \sim \pi_\tasks(\cdot | s)}\left[ Q_\tasks(s, a) - \alpha \log \pi_\tasks(a | s) \right],
\end{equation}
where $\pi_\text{int}$ refers to the set of intentions $\{ \pi_{\tasks_{\text{main}}}, \pi_{\tasks_{\text{1}}}, \dots, \pi_{\tasks_{\text{K}}} \}$ and $\buffer_\text{all}$ refers to buffer containing every transition from interactions and demonstrations, as is done in \cite{vecerikLeveragingDemonstrationsDeep2018,kalashnikovQTOptScalableDeep2018}.

For policy evaluation, the soft Q-functions $Q_\tasks$ for each $\pi_\tasks$ minimize the joint soft Bellman residual
\begin{equation}
    \label{complete_PE}
    \mathcal{L}(Q) = \sum_{\tasks \in \tasks_\text{all}}\mathbb{E}_{(s, a, s') \sim \buffer_\text{all}, a' \sim \pi_\tasks(\cdot | s')}\left[ (Q_\tasks(s, a) - \delta_\tasks)^2 \right],
\end{equation}
\begin{equation}
    \delta_\tasks = R_\tasks(s, a) + \gamma \left( Q_\tasks(s', a') - \alpha \log \pi_\tasks(a' | s') \right).
\end{equation}
\hl{Crucially, because each task shares the common $\mdps, \mdpa, \mdpp, \mdprho$, and $\gamma$, and we are using off-policy learning, all tasks can learn from all data, as in \cite{riedmillerLearningPlayingSolving2018}.}

\subsubsection{The Scheduler} \label{sec:method_scheduler}
SAC-X formulates learning the scheduler by maximizing the expected return of the main task \cite{riedmillerLearningPlayingSolving2018}.
In particular, let $H$ be the number of possible intention switches within an episode and let each chosen intention execute for $\xi$ timesteps.
The $H$ intention choices made within the episode are defined as $\tasks^{0:H-1} = \left\{ \tasks^{(0)}, \dots, \tasks^{(H-1)} \right\}$, where $\tasks^{(h)} \in \tasks_\text{all}$.
The return of the main task, given chosen intentions, is then defined as
\begin{eqnarray}
    G_{\tasks_\text{main}}(\tasks^{0:H-1}) = \sum_{h=0}^{H - 1} \sum_{t=h\xi}^{(h + 1)\xi - 1} \gamma^t R_{\tasks_\text{main}}(s_t, a_t),
\end{eqnarray}
where $a_t \sim \pi_{\tasks^{(h)}}(\cdot | s_t)$ is the action taken at timestep $t$, sampled from the chosen intention $\tasks^{(h)}$ in the $h^\text{th}$ scheduler period. 
The scheduler for the $h^\text{th}$ period $P_S^h$ aims to maximize the expected main task return: $\mathbb{E}\left[G_{\tasks_\text{main}}(\tasks^{h:H-1}) | P_S^h \right]$.
Although SAC-X describes a method to learn the scheduler \cite{riedmillerLearningPlayingSolving2018}, we find that a combination of two simple task-agnostic heuristics performs similarly in practice (see \cref{sec:scheduler_ablations}).

\hl{Specifically, we use a weighted random scheduler (WRS) combined with handcrafted trajectories (HC).
The WRS forms a prior categorical distribution over the set of tasks, with a higher probability mass $p_{\tasks_{\text{main}}}$ for the main task and $\frac{p_{\tasks_{\text{main}}}}{K}$ for all other tasks.
This approach is comparable to the uniform scheduler from \cite{riedmillerLearningPlayingSolving2018}, with a bias towards the main task.
The HC component is a small set of handcrafted trajectories of tasks that are sampled half of the time, forcing the scheduler to explore trajectories that would clearly be beneficial for completing the main task.
The chosen handcrafted trajectories can be found in our code and in our supplementary material.}

\subsection{Breaking Out of Local Maxima with LfGP}

Returning to the discussion in \cref{sec:local_maximum_ail}, resolving the local maximum problem with LfGP is straightforward. 
Suppose we include a \textit{go-right} auxiliary task with $\buffer^E_{\text{go-right}} = \{ (s^1, a^{12}), (s^2, a^{23}), (s^3, a^{34}) \}$.
When the scheduler chooses the go-right intention, the agent does not exploit the $a^{55}$ action because the go-right discriminator learns that $R(s^5, a^{55}) = -1$. 
Since the transitions are stored in the shared buffer that the main intention also samples from, the agent can quickly obtain the correct, optimal value.

\subsection{Expert Data Collection} 
\label{sec:exp_data_collection}

We assume that each $\tasks \in \tasks_\text{all}$ has, for evaluation purposes only, a binary indicator of success.
In single-task imitation learning where this assumption is valid, expert data is typically collected by allowing the expert to control the agent until success conditions are met.
At that point, the environment is reset following $\rho_0$ and collection is repeated for a fixed number of episodes or $(s,a)$ pairs.
We collect our expert data in this way for each $\tasks$ separately.

%% file: sections/4-experiments.tex
\section{Experiments} 
\label{sec:exp}

\begin{figure}[b]
    \vspace{-3mm}
    \centering
    \begin{subfigure}{\linewidth}
        \includegraphics[width=\textwidth]{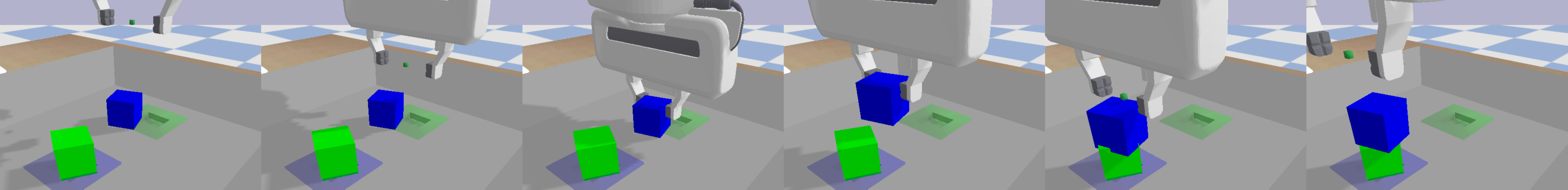}
    \end{subfigure}
    \par \vspace{1mm}
    \begin{subfigure}{\linewidth}
        \includegraphics[width=\textwidth]{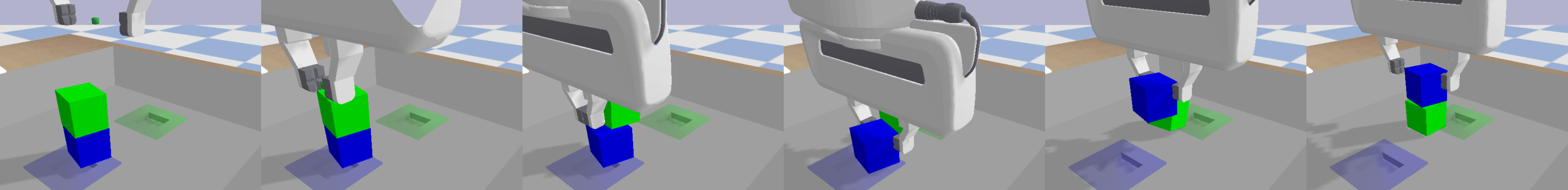}
    \end{subfigure}
    \par \vspace{1mm}
    \begin{subfigure}{\linewidth}
        \includegraphics[width=\textwidth]{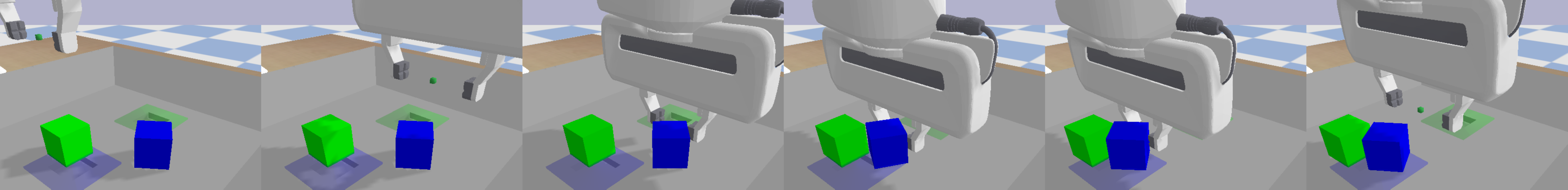}
    \end{subfigure}
    \par \vspace{1mm}
    \begin{subfigure}{\linewidth}
        \includegraphics[width=\textwidth]{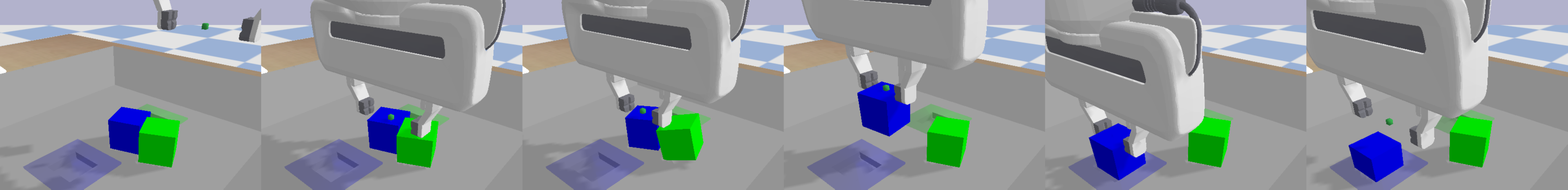}
    \end{subfigure}
    
    \caption{Example successful runs of our four main tasks. Top to bottom: Stack, Unstack-Stack, Bring, Insert. }
    \label{fig:example_runs}
    \vspace{-2mm}
\end{figure}

In this work, we are interested in answering the following questions about LfGP:
\begin{enumerate}
    \item How does the performance of LfGP compare with BC and AIL in challenging manipulation tasks, in terms of success rate and expert sample efficiency?
    \item What parts of LfGP are necessary for success?
    \item How do the policies and action value functions differ between AIL and LfGP?
\end{enumerate}

\begin{figure*}[t]
    \centering
    \includegraphics[width=\textwidth]{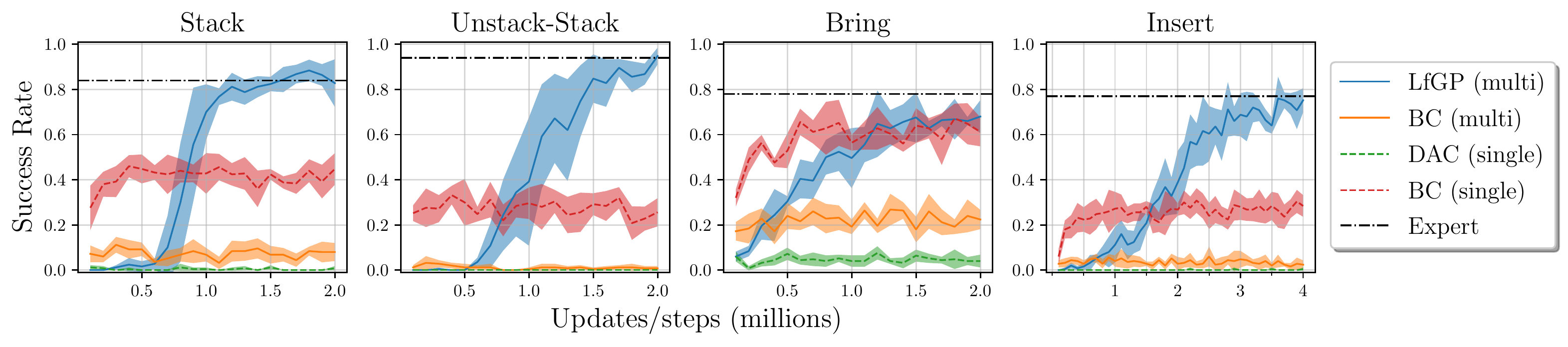}
    \caption{Performance results for LfGP, multitask BC, single-task BC, and DAC on all four tasks considered in this work.
    The $x$-axis corresponds to both gradient updates and environments steps for LfGP and DAC, and gradient updates only for both versions of BC.
    The shaded area corresponds to standard deviation across five seeds. 
    LfGP significantly outperforms the baselines on all tasks, and even in Bring where it is matched by single-task BC, it is far more expert sample efficient. }
    \label{fig:perf_results}
    \vspace{-4mm}
\end{figure*}

\subsection{Experimental Setup} \label{sec:exp_setup}
We complete experiments in a simulation environment containing a Franka Emika Panda manipulator, one green and one blue block in a tray, fixed zones corresponding to the green and blue blocks, and one slot in each zone with $<1$mm tolerance for fitting the blocks (see bottom right of \cref{fig:example_runs}).
The robot is controlled via delta-position commands, and the blocks and end-effector can both be reset anywhere above the tray.
The environment is designed such that several different challenging tasks can be completed within a common observation and action space.
The main tasks that we investigate are Stack, Unstack-Stack, Bring, and Insert (see \cref{fig:example_runs}).
For more details on our environment and definitions of task success, see supplementary material link in \cref{fn:supp}.
We also define a set of auxiliary tasks: Open-Gripper, Close-Gripper, Reach, Lift, Move-Object, and Bring (Bring is both a main task and an auxiliary task for Insert), all of which are reusable between main tasks. 

We compare our method to several standard multitask and single-task baselines.
A multitask algorithm simultaneously learns to complete a main task as well as auxiliary tasks, while the single-task algorithms only learn to complete the main task.
In general, we consider a multitask algorithm to be more useful than a single-task algorithm, given the potential to reuse expert data and trained models for learning new tasks. %
To ensure a fair comparison, we provide single-task algorithms with an equivalent amount of \textit{total} expert data as our multitask methods, as shown in \cref{tab:env_details}.

In our main experiments, we compare LfGP to a multitask variant of behavioural cloning (BC), single-task BC, and Discriminator-Actor-Critic (DAC) \cite{kostrikovDiscriminatorActorCriticAddressingSample2019}, a state-of-the-art approach to AIL.
We train multitask BC with a multitask mean squared error objective, 
\begin{align}
    \mathcal{L}(\pi_{\text{int}}) = \sum_{\tasks \in \tasks_\text{all}} \ \sum_{(s,a) \in \buffer^E_{\tasks}} \left(\pi_{\tasks}(s) - a \right)^2,
\end{align}
while BC is trained with the corresponding single task version.
Following recent trends in improving BC performance, we train our BC baselines with the same number of gradient updates as LfGP and DAC, evaluating the policies at the same frequency.
This adjustment has been shown to dramatically increase the performance of BC \cite{mandlekarWhatMattersLearning2021, hussenotHyperparameterSelectionImitation2021}, particularly compared to the more common practice of using early stopping, as is done in \cite{kostrikovDiscriminatorActorCriticAddressingSample2019, hoGenerativeAdversarialImitation2016}.
We validate that this change significantly improves BC performance in our ablation study (see \cref{sec:baseline_ablations}).

We gather expert data by first training an expert policy using Scheduled Auxiliary Control (SAC-X) \cite{riedmillerLearningPlayingSolving2018}.
We then run the expert policies to collect various amounts of expert data as described in \cref{sec:exp_data_collection} and \cref{tab:env_details}.
We also collect an extra 200 expert $(s_T,\boldsymbol{0})$ pairs per auxiliary task, where $T$ refers to the final timestep of an individual episode and $\boldsymbol{0}$ is an action of all zeros.
This is equivalent to adding example data, as is done in example-based RL \cite{fuVariationalInverseControl2018}.
This addition improved final task performance, likely because it biases the reward towards completing the final task.
It is worth noting that, in the real world, final states are easier to collect than full demonstrations, and LfGP does not require any modifications to accommodate these extra examples.
Finally, even without this addition, LfGP still outperforms the baselines (see \cref{sec:dataset_ablations}).

\begin{table}
	\centering
	\small
	\begin{tabularx}{\columnwidth}{l|llXXX}
		                                                           &     Task & Dataset Sizes & Reuse & Single & Total
		\\\midrule
		\textit{Multi}                                            & Stack & S\textbf{OCRLM}: 1k/task          & \textbf{5k}              & 1k                  & 6k   \\
		\textit{task}                                              & U-Stack & U\textbf{OCRLM}: 1k/task          & \textbf{5k}              & 1k                  & 6k   \\
		                                                           & Bring & \textbf{BOCRLM}: 1k/task          & \textbf{6k}              & 0                   & 6k   \\
		                                                           & Insert    & I\textbf{BOCRLM}: 1k/task         & \textbf{6k}              & 1k                  & 7k   \\ \midrule
		\textit{Single}                                            & Stack      & S: 6k        & 0                & 6k                  & 6k   \\
		\textit{Task}                                              & U-Stack     & U: 6k         & 0                & 6k                 & 6k   \\
		                                                           & Bring     & B: 6k         & 0                & 6k                 & 6k   \\
		                                                           & Insert    & I: 6k         & 0                & 7k                 & 7k   \\
	\end{tabularx}
	\caption{
            The number of $(s,a)$ pairs used for each main and auxiliary task.
            The table illustrates the reusability of the expert data used to generate the performance results described in \cref{sec:perf_res}. 
	    Each letter under ``Dataset Sizes'' is the first letter of a single (auxiliary) task, and bolded letters indicate that a dataset was reused for more        than one main task (e.g., \textbf{O}pen-Gripper was used for all four main tasks). 
	    Multitask methods (e.g., LfGP) are able to reuse a large portion of the expert data, while single-task methods (e.g., single-task BC) cannot.
        }
	\label{tab:env_details}
	\vspace{-4mm}
\end{table}

\subsection{Performance Results} \label{sec:perf_res}
\begin{figure*}[t]
    \centering
    \includegraphics[width=\textwidth]{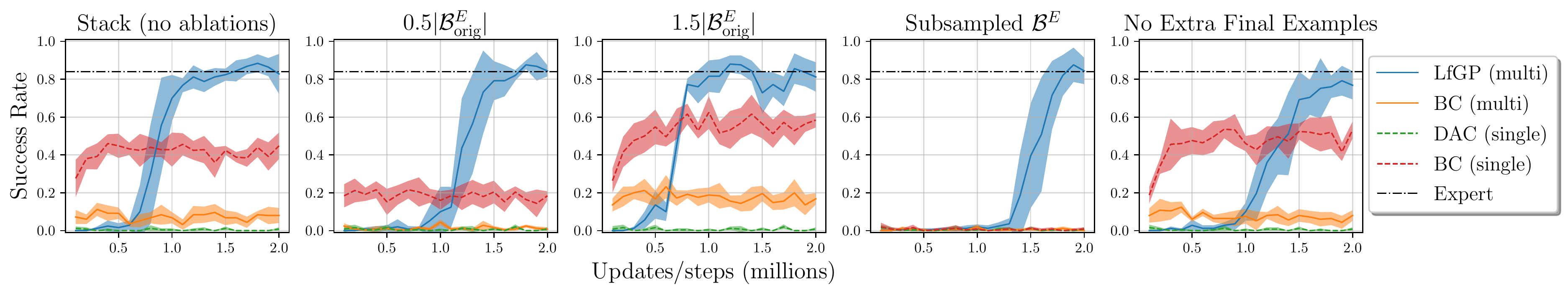}
    \caption{
        Various dataset ablations for LfGP and all baselines, including dataset size, subsampling of expert dataset, and replacement of extra $(s_T,\boldsymbol{0})$ pairs with an equivalent amount of regular trajectory $(s, a)$ pairs.
        In all cases, LfGP still significantly outperforms all baselines.
    }
    \label{fig:dataset_ablations}
    \vspace{-4mm}
\end{figure*}

\begin{figure*}[t]
    \centering
    \begin{subfigure}{0.31\textwidth}
        \includegraphics[width=\textwidth]{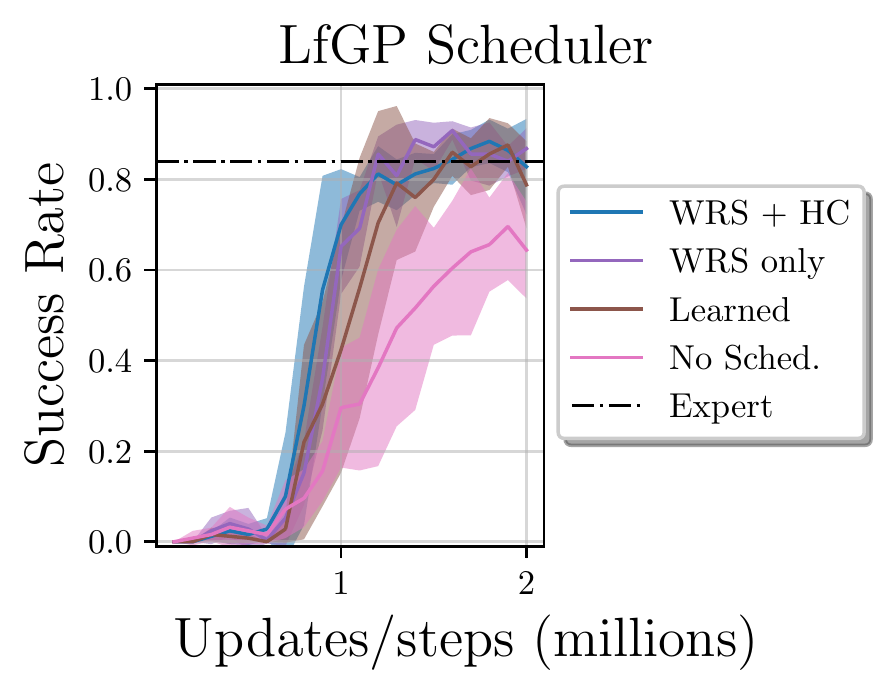}
    \end{subfigure}
    \begin{subfigure}{0.31\textwidth}
        \includegraphics[width=\textwidth]{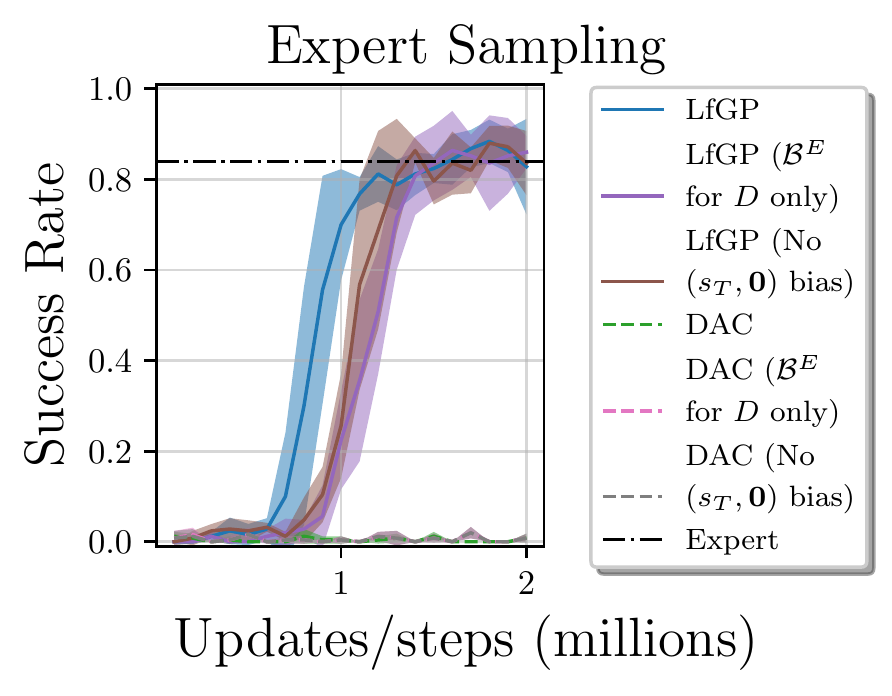}
    \end{subfigure}
    \begin{subfigure}{0.31\textwidth}
        \includegraphics[width=\textwidth]{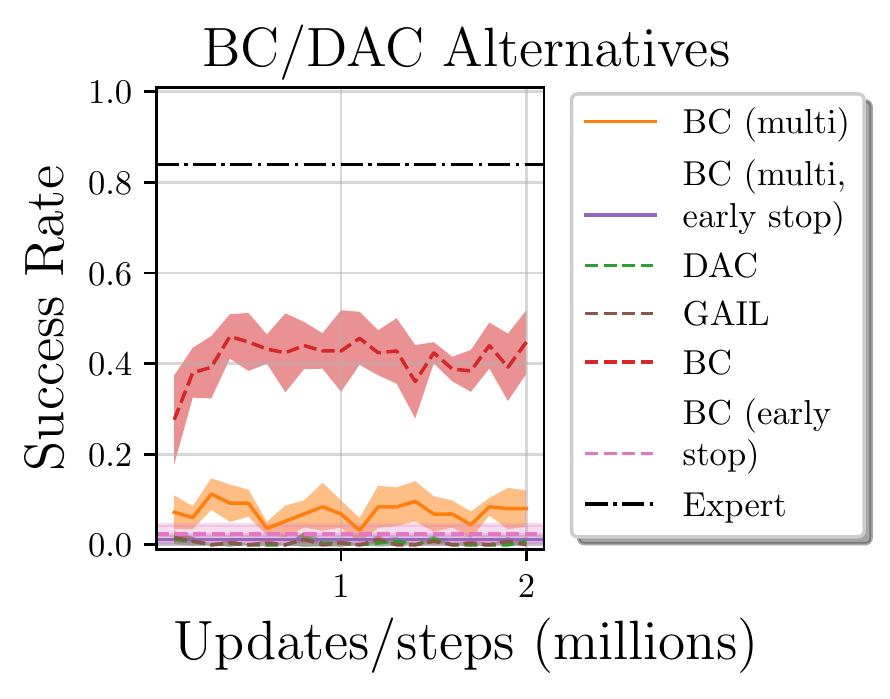}
    \end{subfigure}
    \caption{
        Left: Scheduler ablations for training LfGP, WRS is weighted random scheduler, HC is handcraft; Middle: Expert sampling ablations for training LfGP/DAC; Right: Baseline ablations for training BC/DAC.
    }
    \label{fig:scheduler_sampling_baseline_ablations}
    \vspace{-3mm}
\end{figure*}

Performance results for all methods and main tasks are shown in \cref{fig:perf_results}.
We freeze the policies every 100k steps and evaluate those policies for 50 randomized episodes, using only the mean action outputs for stochastic policies.
For all algorithms, we test across five seeds and report the mean and standard deviation of all seeds.

In Stack, Unstack-Stack, and Insert, LfGP achieves expert performance, while the baselines all perform significantly worse.
In Bring, LfGP does not quite achieve expert performance, and is matched by single-task BC. 
However, we note that LfGP is much more expert data efficient than single-task BC because it reuses auxiliary task data (see \cref{tab:env_details}).
A more direct comparison is multitask BC, which performs much more poorly than LfGP across all tasks, including Bring.
Intriguingly, DAC also performs very poorly on all tasks, a phenomenon that we further explore in \cref{sec:learned_model_analysis}.

\subsection{Ablation Study}
\label{sec:ablation}

While the fundamental idea of LfGP is relatively straightforward, it is worth considering alternatives to some of the specific choices made for our experiments.
In this section, we complete an ablation study where we vary (a) the expert dataset, including size, subsampling, and inclusion of extra $(s_T,\boldsymbol{0})$ pairs; (b) the type of scheduler used for LfGP (see \cref{sec:method_scheduler}); (c) the sampling strategy used for expert data; and (d) the method for training our baselines.
To reduce the computational load of completing these experiments, all of these variations were carried out exclusively for our Stack task.
All ablation results are shown in \cref{fig:dataset_ablations} and \cref{fig:scheduler_sampling_baseline_ablations}.

\subsubsection{Dataset Ablations} \label{sec:dataset_ablations}
We tested the following dataset variations: (a) half and one and a half times the original expert dataset size; (b) subsampling $\buffer^E$, taking only every 20th timestep, as is done in \cite{kostrikovDiscriminatorActorCriticAddressingSample2019, hoGenerativeAdversarialImitation2016}; and (c) replacing the 200 extra $(s_T,\boldsymbol{0})$ pairs in each buffer with 200 regular trajectory $(s, a)$ pairs.
Notably, even in the challenging regimes of halving and subsampling the dataset, LfGP still learns an expert-level policy (albeit more slowly).

\subsubsection{Scheduler Ablations} \label{sec:scheduler_ablations}

\begin{figure*}[ht]
    \centering
    \includegraphics[width=\linewidth]{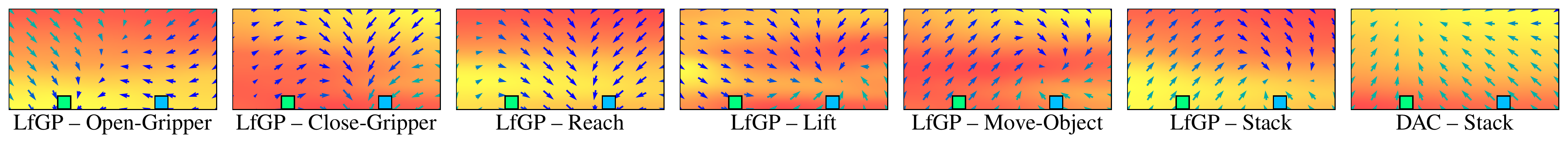}
    \caption{
        The policy outputs (arrows) and Q values (background) for each LfGP task and for DAC at 200k environment steps. 
        The arrows show velocity direction/magnitude, blue $\rightarrow$ green indicates open-gripper $\rightarrow$ close-gripper. 
        For Q values, red $\rightarrow$ yellow indicates low $\rightarrow$ high.
        The LfGP policies and Q functions are reasonable for all tasks, while DAC has only learned to reach toward and above the green block.
    }
    \label{fig:lfgp_dac_q}
    \vspace{-4mm}
\end{figure*}

We tested the following scheduler variations: (a) Weighted Random Scheduler (WRS) only, removing the Handcrafted (HC) addition; (b) a learned scheduler, as is used in \cite{riedmillerLearningPlayingSolving2018}; and (c) no scheduler, in which only the main task is attempted, akin to the Intentional Unintentional Agent \cite{riedmillerLearningPlayingSolving2018, cabiIntentionalUnintentionalAgent2017}.
Both WRS versions learn slightly faster than the learned scheduler, but all three methods outperform the No Scheduler ablation, replicating results from \cite{riedmillerLearningPlayingSolving2018} demonstrating the importance of actually exploring all auxiliary tasks.
Perhaps surprisingly, the HC modification made little difference compared with WRS only, but it is possible that for even more complex tasks, this could change.

\subsubsection{Expert Sampling Ablations} \label{sec:expert_sampling_ablations}
For our main performance experiments, we modified standard AIL in two ways: (a) we added expert buffer sampling to $\pi$ and $Q$ updates, in addition to the $D$ updates, as is done in \cite{vecerikLeveragingDemonstrationsDeep2018,kalashnikovQTOptScalableDeep2018}; and (b) we biased the sampling of $\buffer^E$ when training $D$ to be $95\%$ final $(s_T,\boldsymbol{0})$ pairs.
We tested both LfGP and DAC without these additions.
For LfGP, although these modifications improve learning speed, they are not required to generate an expert policy.
For DAC, performance is quite poor regardless of these adjustments.

\subsubsection{Baseline Ablations} \label{sec:baseline_ablations}

To verify that we evaluated against fair baselines, we tested two alternatives to those used for our main performance experiments: (a) an early stopping variation of BC, in which each expert buffer is divided into a 70\%/30\% train/validation split, taking the policy after validation error has not improved for 100 epochs; and (b) the on-policy variant of DAC, also known as Generative Adversarial Imitation Learning (GAIL) \cite{hoGenerativeAdversarialImitation2016}.
Notably, the early stopping variants of BC, commonly used as baselines in other AIL work \cite{hoGenerativeAdversarialImitation2016, kostrikovDiscriminatorActorCriticAddressingSample2019, zolnaTaskRelevantAdversarialImitation2021} perform dramatically more poorly than those used in our experiments, verifying recent trends \cite{mandlekarWhatMattersLearning2021, hussenotHyperparameterSelectionImitation2021}.

%% file: sections/5-analysis.tex
\section{Learned Model Analysis} \label{sec:learned_model_analysis}
\begin{figure}[b]
    \vspace{-4mm}
    \centering
    \includegraphics[width=\linewidth]{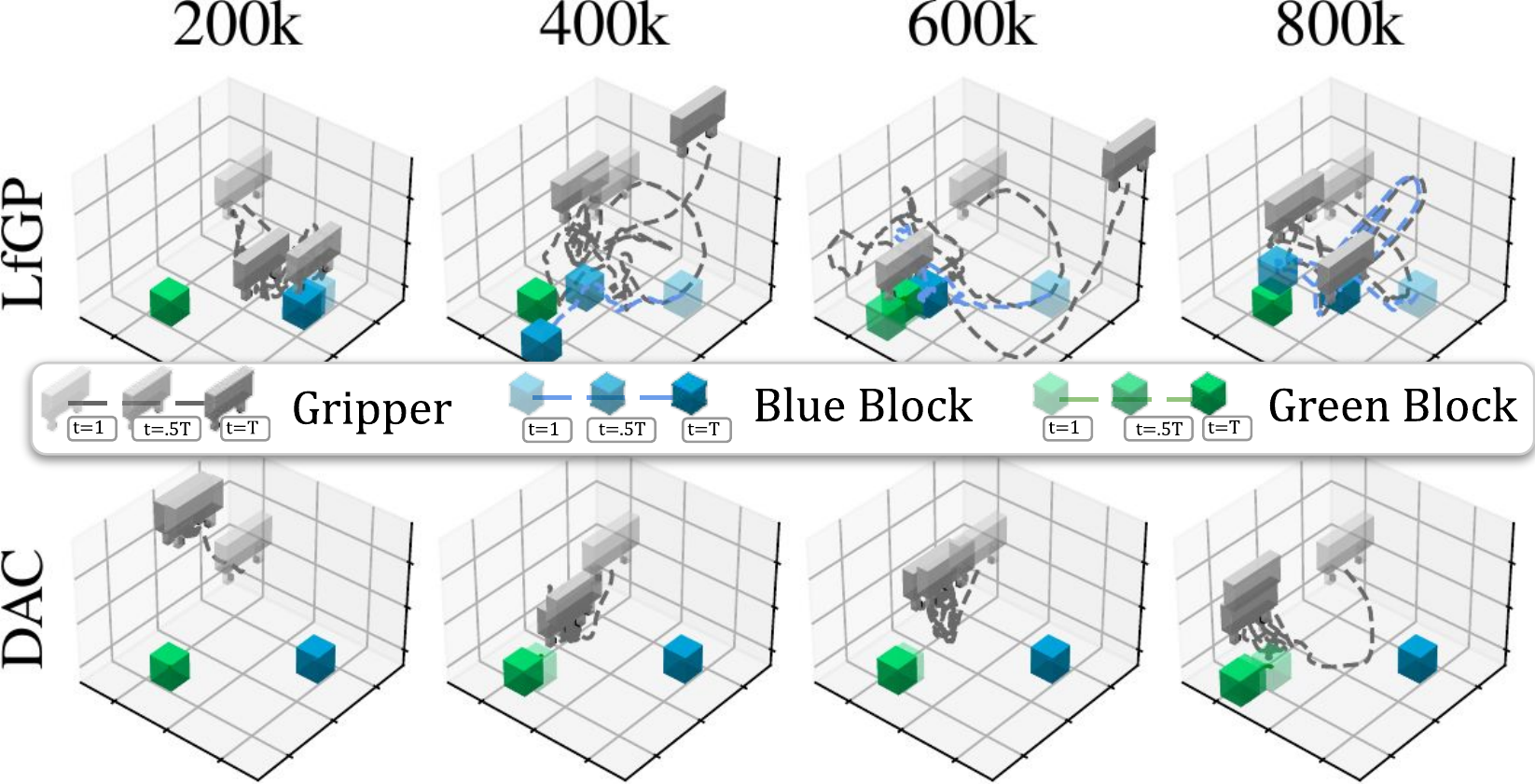}
    \caption{
        LfGP and DAC trajectories of the gripper, blue block, and green block for four stack episodes with consistent initial conditions throughout the learning process. 
        The LfGP episodes, each including auxiliary task sub-trajectories, demonstrate significantly more variety than the DAC trajectories.
    }
    \label{fig:lfgp_dac_snapshots}
\end{figure}

In this section, we further examine the learned Stack models of LfGP and DAC.
We take snapshots of the average performing models from LfGP and DAC at four points during learning: 200k, 400k, 600k, and 800k model updates and environment steps.
Although the initial gripper and block positions are randomized between episodes during learning, for each snapshot, we reset the stacking environment to a single set of representative initial conditions.
We then run the snapshot policies for a single exploratory trajectory, using the stochastic outputs of each policy as well as, for LfGP, the WRS+HC scheduler.
Trajectories from these runs are shown in \cref{fig:lfgp_dac_snapshots}.

DAC is unable to learn to grasp or even reach the blue block and ultimately settles on a policy that learns to reach and hover near the green block.
This is understandable---DAC learns a deceptive reward for hovering above the green block regardless of the position of the blue block, because it has not sufficiently explored the alternative of first grasping the blue block.
Even if hovering above the green block does not fully match the expert data, the DAC policy receives some reward for doing so, as evidenced by the learned Q value on the right side of \cref{fig:lfgp_dac_q}.

In comparison, even after only 200k environment steps, LfGP learns to reach and push the blue block, and by 600k steps, grasp, move, and nearly stack it.
By enforcing exploration of sub-tasks that are crucial to completing the main task, LfGP ensures that the distribution of expert stacking data is fully matched.

%% file: sections/2-background.tex
\section{Related Work}

Imitation learning is often divided into two main categories: behavioural cloning (BC) \cite{rossReductionImitationLearning2011, ablettSeeingAllAngles2021} and inverse reinforcement learning (IRL) \cite{ngAlgorithmsInverseReinforcement2000, abbeelApprenticeshipLearningInverse2004}. 
BC recovers the expert policy via supervised learning, but it suffers from compounding errors due to covariate shift \cite{rossReductionImitationLearning2011, ablettFightingFailuresFIRE2020}.
Alternatively, IRL partially alleviates the covariate shift problem by estimating the reward function and then applying RL using the learned reward. 
A popular approach to IRL is adversarial imitation learning (AIL) \cite{hoGenerativeAdversarialImitation2016, kostrikovDiscriminatorActorCriticAddressingSample2019, hausmanMultiModalImitationLearning2017}, in which the expert policy is recovered by matching the occupancy measure between the generated data and the demonstration data. 
Our proposed method enhances existing AIL algorithms by enabling exploration of key auxiliary tasks via the use of a scheduled multitask model, simultaneously resolving the susceptibility of AIL to deceptive rewards.

Agents learned via hierarchical reinforcement learning (HRL), which act over multiple levels of temporal abstractions in long planning horizon tasks, are shown to provide more effective exploration than agents operating over only a single level of abstraction \cite{riedmillerLearningPlayingSolving2018, suttonMdpsSemiMdpsFramework1999, nachumWhyDoesHierarchy2019}.
Our approach for learning agents most closely resembles hierarchical AIL methods that attempt to combine AIL with HRL \cite{hausmanMultiModalImitationLearning2017, hendersonOptionGANLearningJoint2018, sharmaDirectedInfoGAILLearning2019, jingAdversarialOptionAwareHierarchical2021}. 
Existing work \cite{hendersonOptionGANLearningJoint2018, sharmaDirectedInfoGAILLearning2019, jingAdversarialOptionAwareHierarchical2021} often formulates the hierarchical agent using the Options framework \cite{suttonMdpsSemiMdpsFramework1999} and learns the reward function with AIL \cite{hoGenerativeAdversarialImitation2016}. 
Both \cite{hendersonOptionGANLearningJoint2018} and \cite{jingAdversarialOptionAwareHierarchical2021} leverage task-specific expert demonstrations to learn options using mixture-of-experts and expectation-maximization strategies, respectively. 
In contrast, our work focuses on expert demonstrations that include multiple reusable auxiliary tasks, each of which has clear semantic meaning.

In the multitask setting, \cite{hausmanMultiModalImitationLearning2017} and \cite{sharmaDirectedInfoGAILLearning2019} leverage unsegmented, multitask expert demonstrations to learn low-level policies via a latent variable model.
Other work has used a large corpus of unsegmented but semantically meaningful ``play" expert data to bootstrap policy learning \cite{lynchLearningLatentPlans2019, guptaRelayPolicyLearning2019}.
We define our expert dataset as being derived from \textit{guided} play, in that the expert completes semantically meaningful auxiliary tasks with provided transitions, reducing the burden on the expert to generate these data arbitrarily and simultaneously providing auxiliary task labels. 
Compared with learning from unsegmented demonstrations, the use of segmented demonstrations, as in \cite{codevillaEndtoEndDrivingConditional2018}, ensures that we know which auxiliary tasks our model will be learning, and opens up the possibility of expert data reuse and also transfer learning.
Finally, we deviate from the Options framework and build upon Scheduled Auxiliary Control (SAC-X) to train our hierarchical agent, since SAC-X has been shown to work well for challenging manipulation tasks \cite{riedmillerLearningPlayingSolving2018}.

%% file: sections/6-conclusion.tex
\section{Limitations}
Our approach is not without limitations.
While we were able to use LfGP in six and seven-task settings, the number of tasks for which this method would become intractable is unclear.
LfGP needs access to segmented expert data as well; in many cases, this is reasonable, and is also required to be able to reuse auxiliary task data between main tasks, but it does necessitate extra care during expert data collection.
Also, LfGP requires pre-defined auxiliary tasks: while this is a common approach to hierarchical RL (see \cite{pateriaHierarchicalReinforcementLearning2021}, Section 3.1, for numerous examples), choosing these tasks may sometimes present a challenge.
Finally, compared with methods that use offline data exclusively (e.g., BC), for our tasks, LfGP requires many online environment steps to learn a high-quality policy. 
This data gathering could be costly if human supervision was necessary.
It is worth noting that, because LfGP is already a multitask method, this final point could be partially resolved through the use of multitask reset-free RL \cite{guptaResetFreeReinforcementLearning2021}.

\section{Conclusion}
We have shown how adversarial imitation learning can fail at challenging manipulation tasks because it learns deceptive rewards.
We demonstrated that this can be resolved with Learning from Guided Play (LfGP), in which we introduce auxiliary tasks and the corresponding expert data, \textit{guiding} the agent to \textit{playfully} explore parts of the state and action space that would have been avoided otherwise.
We demonstrated that our method dramatically outperforms both BC and AIL baselines, particularly in the case of AIL.
Furthermore, our method can leverage reusable expert data, making it significantly more expert sample efficient than the highest-performing baseline, and its learned auxiliary task models can be applied to transfer learning.
In future work, we intend to investigate transfer learning to determine if overall policy learning time can be reduced.

%% file: sections/acknowledgements.tex
\section*{Acknowledgements}
We gratefully acknowledge the Digital Research Alliance of Canada and NVIDIA Inc., who provided the GPUs used in this work through their Resources for Research Groups Program and their Hardware Grant Program, respectively.

%% file: sections/appendix.tex
\appendices

\section{Learning from Guided Play Algorithm}
The complete pseudo-code is given in \cref{alg:DACX}. Our implementation builds on RL Sandbox \citeapp{rl_sandbox}, an open-source PyTorch \citeapp{paszkePyTorchImperativeStyle2019} framework for RL algorithms. For learning the discriminators, we follow DAC and apply a gradient penalty for regularization \citeapp{gulrajaniImprovedTrainingWasserstein2017, kostrikovDiscriminatorActorCriticAddressingSample2019}. We optimize the intentions via the reparameterization trick \citeapp{kingmaAutoEncodingVariationalBayes2013}. As is commonly done in deep RL, we use the Clipped Double Q-Learning trick \citeapp{fujimotoAddressingFunctionApproximation2018} to mitigate overestimation bias \citeapp{vanhasseltDeepReinforcementLearning2016} and use a target network to mitigate learning instability \citeapp{mnihHumanlevelControlDeep2015} when training the policies and Q-functions. We also learn the temperature parameter $\alpha_{\tasks}$ separately for each task $\tasks$ (see Section 5 of \citeapp{haarnojaSoftActorCriticAlgorithms2019} for more details on learning $\alpha$). For Generative Adversarial Imitation Learning (GAIL), we use a common open-source PyTorch implementation \citeapp{kostrikovPyTorchImplementationsReinforcement2018}. The hyperparameters chosen for all methods are provided in \cref{sec:hyperparameters}.
Please see videos at \url{papers.starslab.ca/lfgp} for examples of what LfGP looks like in practice.

\begin{algorithm}[htb]
\caption{Learning from Guided Play (LfGP)}
\label{alg:DACX}
\textbf{Input}: Expert replay buffers $\buffer^E_{\text{main}}, \buffer^E_{1}, \dots, \buffer^E_{K}$, scheduler period $\xi$, sample batch size $N$\\
\textbf{Parameters}: Intentions $\pi_\tasks$ with corresponding Q-functions $Q_\tasks$ and discriminators $D_\tasks$, and scheduler $\pi_S$ (e.g. with Q-table $Q_S$)
\begin{algorithmic}[1] %
\STATE Initialize replay buffer $\mathcal{B}$ \\
\FOR{$t = 1, \dots,$}
    \STATE \# Interact with environment
    \STATE For every $\xi$ steps, select intention $\pi_\tasks$ using $\pi_S$
    \STATE Select action $a_t$ using $\pi_\tasks$
    \STATE Execute action $a_t$ and observe next state $s'_t$
    \STATE Store transition $\langle s_t, a_t, s'_t \rangle$ in $\mathcal{B}$
    \STATE 
    \STATE \# Update discriminator $D_{\tasks'}$ for each task $\tasks'$
    \STATE Sample $\left\{ (s_i, a_i) \right\}_{i=1}^{N} \sim \mathcal{B}$
    \FOR{each task $\tasks'$}
        \STATE Sample $\left\{ (s'_i, a'_i) \right\}_{i=1}^{B} \sim \buffer^E_k$
        \STATE Update $D_{\tasks'}$ following Eq. (1) using GAN + Gradient Penalty
    \ENDFOR
    \STATE
    \STATE \# Update intentions $\pi_{\tasks'}$ and Q-functions $Q_{\tasks'}$ for each task $\tasks'$
    \STATE Sample $\left\{ (s_i, a_i) \right\}_{i=1}^{N} \sim \mathcal{B}$
    \STATE Compute reward $D_{\tasks'}(s_i, a_i)$ for each task $\tasks'$
    \STATE Update $\pi$ and $Q$ following Eq. (4) and Eq. (5)  %
    \STATE
    \STATE \# \textit{Optional }Update learned scheduler $\pi_S$
    \IF{at the end of effective horizon}
        \STATE Compute main task return $G_{\tasks_\text{main}}$ using reward estimate from $D_\text{main}$
        \STATE Update $\pi_S$ (e.g. update Q-table $Q_S$ following \cref{eqn:EMA} and recompute Boltzmann distribution)
    \ENDIF
\ENDFOR
\end{algorithmic}
\vspace{2mm}
\end{algorithm}

\subsection{Scheduler Details}
\subsubsection{Learning the Scheduler}
As stated in our paper, our main experiments used a simple weighted random scheduler with handcrafted trajectories.
In this section, we provide the details of our learned scheduler.
Following \citeapp{riedmillerLearningPlayingSolving2018}, let $H$ be the total number of possible intention switches within an episode and let each chosen intention execute for $\xi$ timesteps.
The $H$ intention choices made within the episode are defined as $\tasks^{0:H-1} = \left\{ \tasks^{(0)}, \dots, \tasks^{(H-1)} \right\}$, where $\tasks^{(h)} \in \tasks_\text{all}$.
The main task's return given chosen intentions is then defined as
\begin{align} \tag{A.1}
    G_{\tasks_\text{main}}(\tasks^{0:H-1}) = \sum_{h=0}^{H - 1} \sum_{t=h\xi}^{(h + 1)\xi - 1} \gamma^t R_{\tasks_\text{main}}(s_t, a_t),
\end{align}
where $a_t \sim \pi_{\tasks^{(h)}}(\cdot | s_t)$ is the action taken at timestep $t$, sampled from the chosen intention $\tasks^{(h)}$ in the $h^\text{th}$ scheduler period. 
We further define the Q-function for the scheduler as $Q_S(\tasks^{0:h-1}, \tasks^{(h)}) = \mathbb{E}_{\hl{\tasks^{h:H-1} \sim P_S^{h:H-1}}}\left[ G_{\tasks_\text{main}}(\tasks^{h:H-1}) | \tasks^{0:h-1} \right]$ and represent the scheduler for the $h^\text{th}$ period as a \hl{softmax} distribution \hl{$P_S^h$} over \hl{$\{ Q_S(\tasks^{0:h-1}, \tasks_{\text{main}}), Q_S(\tasks^{0:h-1}, \tasks_1), \dots, Q_S(\tasks^{0:h-1}, \tasks_K) \} $}.
The scheduler maximizes the expected return of the main task following the scheduler:
\begin{align} \tag{A.2}
    \mathcal{L}(S) = \mathbb{E}_{\hl{\tasks^{(0)} \sim P_S^0}}\left[ Q_S(\emptyset, \tasks^{(0)}) \right].
    \label{scheduler}
\end{align}
We use Monte Carlo returns to estimate $Q_S$, estimating the expected return using the exponential moving average:

\begin{equation} \tag{A.3}
\begin{aligned}
    \label{eqn:EMA}
    Q_S(\tasks^{0:h-1}, \tasks^{(h)}) = (1 - \phi)Q_S(\tasks^{0:h-1}, \tasks^{(h)})\\ + \phi\,G_{\tasks_\text{main}}(\tasks^{h:H}),
\end{aligned}
\end{equation}
where $\phi \in [0, 1]$ represents the amount of discounting on older returns and $G_{\tasks_\text{main}}(\tasks^{h:H})$ is the cumulative discounted return of the trajectory starting at timestep $h\xi$.

\subsection{Weighted Random Scheduler Plus Handcrafted Trajectories}
As stated in our paper, the main experiments were completed with the described weighted random scheduler (WRS) combined with some simple handcrafted trajectories (HC) that we expected to be beneficial for learning each of the main tasks.
In this section, we provide further details of these handcrafted scheduler trajectories.
Given a chosen proportion hyperparameter (0.5 in our experiments), we randomly sampled full trajectories from the lists below at the beginning of training episodes, and otherwise sampled from the regular WRS.
For all four tasks $\text{Main} = \{ \text{Stack, Unstack-Stack, Bring, Insert} \} $, we provided the following set of trajectories:
\begin{enumerate}
    \item Reach, Lift, Main, Open-Gripper, Reach, Lift, Main, Open-Gripper.
    \item Reach, Lift, Move-Object, Main, Open-Gripper, Reach, Lift, Move-Object.
    \item Lift, Main, Open-Gripper, Lift, Main, Open-Gripper, Lift, Main.
    \item Main, Open-Gripper, Main, Open-Gripper, Main, Open-Gripper, Main, Open-Gripper.
    \item Move-Object, Main, Open-Gripper, Move-Object, Main, Open-Gripper, Move-Object, Main.
\end{enumerate}
For insert, in addition to the trajectories listed above, we added two more trajectories to specifically accommodate Bring as an auxiliary task:
\begin{enumerate}
    \item Bring, Insert, Open-Gripper, Bring, Insert, Open-Gripper, Bring, Insert.
    \item Reach, Lift, Bring, Insert, Open-Gripper, Reach, Lift, Bring.
\end{enumerate}

\section{Environment Details} \label{sec:env_details}
\begin{figure}[ht]
	\centering
	\includegraphics[width=0.48\textwidth]{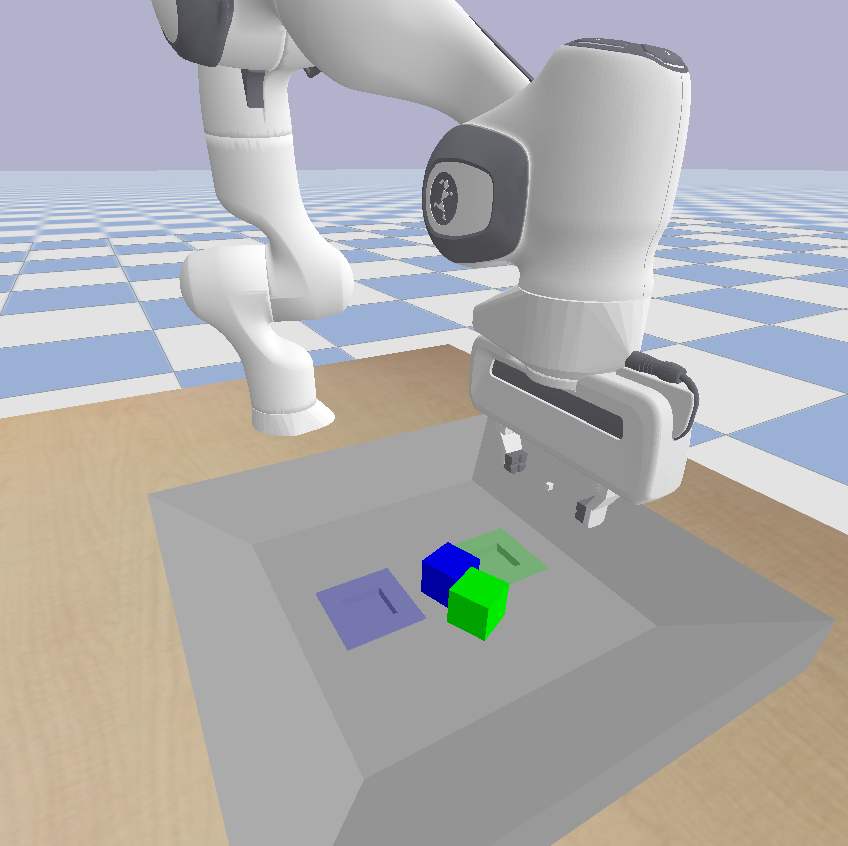}
	\caption{An image of our multitask environment immediately after a reset has been carried out.}
	\label{fig:env_example}
\end{figure}
\begin{table}
	\centering
	\small
    \caption{The components used in our environment observations, common to all tasks. Grip finger position is a continuous value from 0 (closed) to 1 (open).}
	\begin{tabular}{lrlll}
	    \toprule
		    Component  & Dim            & Unit       & Privileged?  &  Extra info                        \\
		\midrule
		EE pos.                & 3              & m          & No           & rel. to base           \\
		EE velocity            & 3              & m/s        & No           & rel. to base           \\
		Grip finger pos.       & 6              & [0, 1]       & No           & current, last 2        \\
		Block pos.            & 6              & m          & Yes            & both blocks           \\
		Block rot.            & 8              & quat       & Yes            & both blocks           \\
		Block trans vel.      & 6              & m/s        & Yes            & rel. to base           \\
		Block rot vel.        & 6              & rad/s      & Yes            & rel. to base           \\
		Block rel to EE       & 6              & m          & Yes            & both blocks           \\
		Block rel to block    & 3              & m          & Yes            & in base frame           \\
		Block rel to slot     & 6              & m          & Yes            & both blocks           \\
		Force-torque          & 6              & N,Nm       & No            & at wrist           \\
		\midrule
		\textbf{Total}        & \textbf{59}      &       &             &         \\
	\end{tabular}
	\label{tab:obs_details}
	\vspace{-5mm}
\end{table}

A screenshot of our environment, simulated in PyBullet \citeapp{coumans2019}, is shown in \cref{fig:env_example}.
We chose this environment because we desired tasks that a) have a large distribution of possible initial states, representative of manipulation in the real world, b) have a shared observation/action space with several other tasks, allowing the use of auxiliary tasks and transfer learning, and c) require a reasonably long horizon and significant use of contact to solve.
The environment contains a tray with sloped edges (to keep the blocks within the reachable workspace of the end-effector), as well as a green and a blue block, each of which is 4 cm $\times$ 4 cm $\times$ 4 cm and has a mass of 100 g.
The dimensions of the lower part of the tray, before reaching the sloped edges, are 30 cm $\times$ 30 cm.
The dimensions of the `bring' boundaries (shaded blue and green regions) are 8 cm $\times$ 8 cm, while the dimensions of the insertion slots, which are directly in the center of each shaded region, are 4.1 cm $\times$ 4.1 cm $\times$ 1 cm.
The boundaries for end-effector movement, relative to the tool center point that is directly between the gripper fingers, are a 30 cm $\times$ 30 cm $\times$ 14.5 cm box, where the bottom boundary is low enough to allow the gripper to interact with objects, but not to collide with the bottom of the tray.

See \cref{tab:obs_details} for a summary of our environment observations.
In this work, we use privileged state information (e.g., block poses), but adapting our method to exclusively use image-based data is straightforward since we do not use hand-crafted reward functions as in \citeapp{riedmillerLearningPlayingSolving2018}.

The environment movement actions are 3-DOF translational position changes, where the position change is relative to the current end-effector position. We leverage PyBullet's built-in position-based inverse kinematics function to generate joint commands.
Our actions also contain a fourth dimension that corresponds to actuating the gripper.
To allow for the use of policy models with exclusively continuous outputs, this dimension accepts any real number, with any value greater than 0 commanding the gripper to open, and any number less than 0 commanding it to close.
Actions are supplied at a rate of 20 Hz, and each training episode is limited to 18 seconds, corresponding to 360 time steps per episode.
For play-based expert data collection, we also reset the environment manually every 360 time steps.
Between episodes, block positions are randomized to any pose within the tray, and the end-effector is randomized to any position between 5 and 14.5 cm above the tray, within the earlier stated end-effector bounds, with the gripper fully opened.
The only exception to these initial conditions is during expert data collection and agent training of the Unstack-Stack task: in this case, the green block is manually set to be on top of the blue block at the start of the episode.

\section{Performance Results for Auxiliary Tasks}
\begin{figure*}[t]
	\centering
	\includegraphics[width=\textwidth]{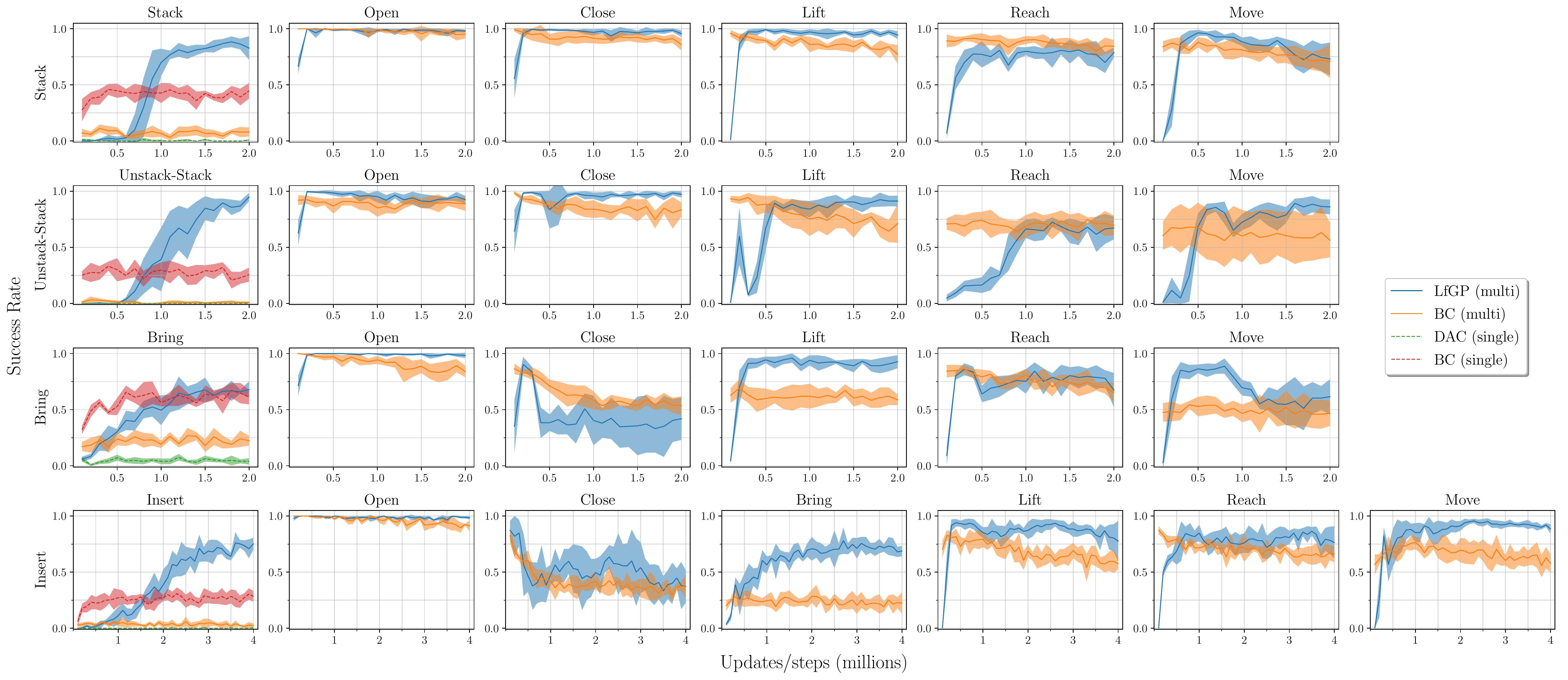}
	\caption{Performance for LfGP and the multitask baselines across all tasks, shaded area corresponds to standard deviation.}
	\label{fig:multi_perf_results}
	\vspace{-5mm}
\end{figure*}
The performance results for all multitask methods and all auxiliary tasks are shown in  \cref{fig:multi_perf_results}.
Multitask BC has gradually decreasing performance on many of the auxiliary tasks as the number of updates increases, which is consistent with mild overfitting.
Intriguingly, however, multitask BC does achieve quite reasonable performance on many of the auxiliary tasks (such as Lift) without needing any of the extra environment interactions required by an online method such as LfGP or DAC.
An interesting direction for future work is to determine whether pretraining via multitask BC could provide any improvements in environment sample efficiency.
We did attempt to do this, but found that it resulted in poorer final performance than training from scratch.

\section{Procedure for Obtaining Experts} \label{sec:expert_gen}
As stated, we used SAC-X \citeapp{riedmillerLearningPlayingSolving2018} to train models that we used for generating expert data.
We used the same hyperparameters that we used for LfGP (see \cref{tab:hyperparameters_ail}), apart from the discriminator, which, of course, does not exist in SAC-X.
See \cref{sec:evaluation} for details on the hand-crafted rewards that we used for training these models.
For an example of gathering play-based expert data, please see our attached video.

We made two modifications to regular SAC-X to speed up learning.
First, we pre-trained a Move-Object model before transferring this model to each of our main tasks, as we did in Section 5.3 of our main paper, since we found that SAC-X would plateau when we tried to learn the more challenging tasks from scratch.
The need for this modification demonstrates another noteworthy benefit of LfGP---when training LfGP, main tasks could be learned from scratch, and generally in fewer time steps, than it took to train our experts.
Second, during transfer to the main tasks, we used what we called a conditional weighted scheduler instead of a Q-Table: we defined weights for every combination of tasks, so that the scheduler would pick each task with probability $P(\tasks^{(h)} | \tasks^{(h-1)})$, ensuring that $\forall \tasks' \in \tasks_{\text{all}}, \sum_{\tasks \in \tasks_{\text{all}}} P(\tasks | \tasks') = 1$.
The weights that we used were fairly consistent between main tasks, and can be found in our packaged code.
The conditional weighted scheduler ensured that every task was still explored throughout the learning process, so that we would have high-quality experts for every auxiliary task in addition to the main task.
This scheduler can be considered as a more complex alternative to the weighted random scheduler or the addition with handcrafted trajectories from our main paper, and again shows the flexibility of using a semantically-meaningful multitask policy with a common observation and action space.

\section{Evaluation} \label{sec:evaluation}

As stated in our paper, we evaluated all algorithms by testing the mean output of the main task policy head in our environment and determining a success rate based on 50 randomly selected resets.
These evaluation episodes were run for 360 time steps to match our training process, and if a condition for success was met within that time, they were recorded as a success. 
The rest of this section describes in detail how we evaluated `success' for each of our main and auxiliary tasks.

As previously stated, we trained experts using a modified SAC-X \citeapp{riedmillerLearningPlayingSolving2018} that required us to define a set of reward functions for each task, which we include in this section.
The authors of \citeapp{riedmillerLearningPlayingSolving2018} focused on sparse rewards  but also showed a few experiments in which dense rewards reduced the time to learn adequate policies, so we chose to use dense rewards.
We note that many of these reward functions are particularly complex and required significant manual shaping effort, further motivating the use of an imitation learning scheme like the one presented in our paper.
It is possible that we could have made do with sparse rewards, such as those used in \citeapp{riedmillerLearningPlayingSolving2018}, but our compute resources made this impractical---for example, in \citeapp{riedmillerLearningPlayingSolving2018}, their agent took 5000 episodes $\times$ 36 actors $\times$ 360 time steps $=$ 64.8 M time steps to learn their stacking task, which would have taken over a month of wall clock time on our fastest machine.
To see the specific values used for the rewards and success conditions described in these sections, please review our code.

Unless otherwise stated, each of the success conditions in this section had to be held for 10 time steps, or 0.5 seconds, before being registered as a success.
This choice was made to prevent registering a success when, for example, the blue block slipped off the green block during the Stack task.

\subsection{Common}
For each of these functions, we use the following common labels:
\begin{itemize}
    \item $p_b$: blue block position,
    \item $v_b$: blue block velocity,
    \item $a_b$: blue block acceleration,
    \item $p_g$: green block position,
    \item $p_e$: end-effector tool center point position (TCP),
    \item $p_s$: center of a block pushed into one of the slots,
    \item $g_1$: (scalar) gripper finger 1 position,
    \item $g_2$: (scalar) gripper finger 2 position, and
    \item $a_g$: (scalar) gripper open/close action.
\end{itemize}
A block is flat on the tray when $p_{b,z} = 0$ or $p_{g,z} = 0$.
To further reduce training time for SAC-X experts, all rewards were set to 0 if $\lVert p_b - p_e \rVert > 0.1$ and $\lVert p_g - p_e \rVert > 0.1$ (i.e., the TCP must be within 10 cm of either block).
During training while using the Unstack-Stack variation of our environment, a penalty of -0.1 was added to each reward if $\lVert p_{g,z} \rVert > 0.001$ (i.e., there was a penalty to all rewards if the green block was not flat on the tray).

\subsection{Stack/Unstack-Stack}
The evaluation conditions for Stack and Unstack-Stack are identical, but in our Unstack-Stack experiments, the environment is manually set to have the green block start on top of the blue block.

\subsubsection{Success}
Using internal PyBullet commands, we check to see whether the blue block is in contact with the green block and is \textit{not} in contact with either the tray or the gripper.

\subsubsection{Reward}
We include a term for checking the distance between the blue block and the spot above the the green block, a term for rewarding increasing distance between the block and the TCP once the block is stacked, a term for shaping lifting behaviour, a term to reward closing the gripper when the block is within a tight reaching tolerance, and a term for rewarding the opening the gripper once the block is stacked.

\subsection{Bring/Insert}

We use the same success and reward calculations for Bring and Insert, but for Bring the threshold for success is 3 cm, and for insert, it is 2.5 mm.

\subsubsection{Success}
We check that the distance between $p_b$ and $p_s$ is less than the defined threshold, that the blue block is touching the tray, and that the end-effector is \textit{not} touching the block.
For Insert, the block can only be within 2.5 mm of the insertion target if it is correctly inserted.

\subsubsection{Reward}
We include a term for checking the distance between the $p_b$ and $p_s$ and a term for rewarding increasing distance between $p_b$ and $p_e$ once the blue block is brought/inserted.

\subsection{Open-Gripper/Close-Gripper}
We use the same success and reward calculations for Open-Gripper and Close-Gripper, apart from inverting the condition.

\subsubsection{Success}
For Open-Gripper and Close-Gripper, we check to see if $a_g < 0$ or $a_g > 0$ respectively.

\subsubsection{Reward}
We include a term for checking the action, as we do in the success condition, and also include a shaping term that discourages high magnitudes of the movement action.

\subsection{Lift}
\subsubsection{Success}
We check to see if $p_{b,z} > 0.06$.

\subsubsection{Reward}
We add a dense reward for checking the height of the block, but specifically also check that the gripper positions correspond to being closed around the block, so that the block does not simply get pushed up the edges of the tray.
We also include a shaping term for encouraging the gripper to close when the block is reached.

\subsection{Reach}
\subsubsection{Success}
We check to see if $\lVert p_e - p_b \rVert < 0.015$.

\subsubsection{Reward}
We have a single dense term to check the distance between $p_e$ and $p_b$.

\subsection{Move-Object}
For Move-Object, we changed the required holding time for success to 1 second, or 20 time steps.

\subsubsection{Success}
We check to see if the $v_b > 0.05$ and $a_b < 5$.
The acceleration condition ensures that the arm has learned to move the block by following a smooth trajectory, rather than vigorously shaking it or continuously picking up and dropping it.

\subsubsection{Reward}
We include a velocity term and an acceleration penalty, as in the success condition, but also include a dense bonus for lifting the block.

\section{Return Plots}
\begin{figure*}[ht]
	\centering
	\includegraphics[width=.95\textwidth]{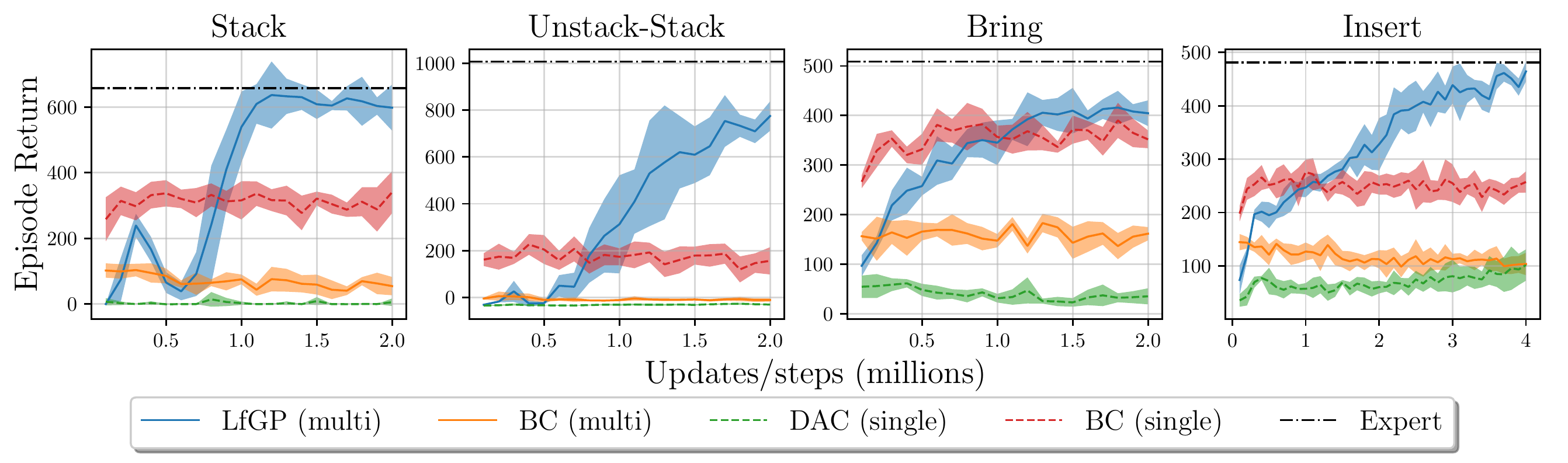}
	\caption{Episode return for LfGP compared with all baselines. Shaded area corresponds to standard deviation.}
	\label{fig:main_return}
\end{figure*}
\begin{figure*}[ht]
	\centering
	\includegraphics[width=.95\textwidth]{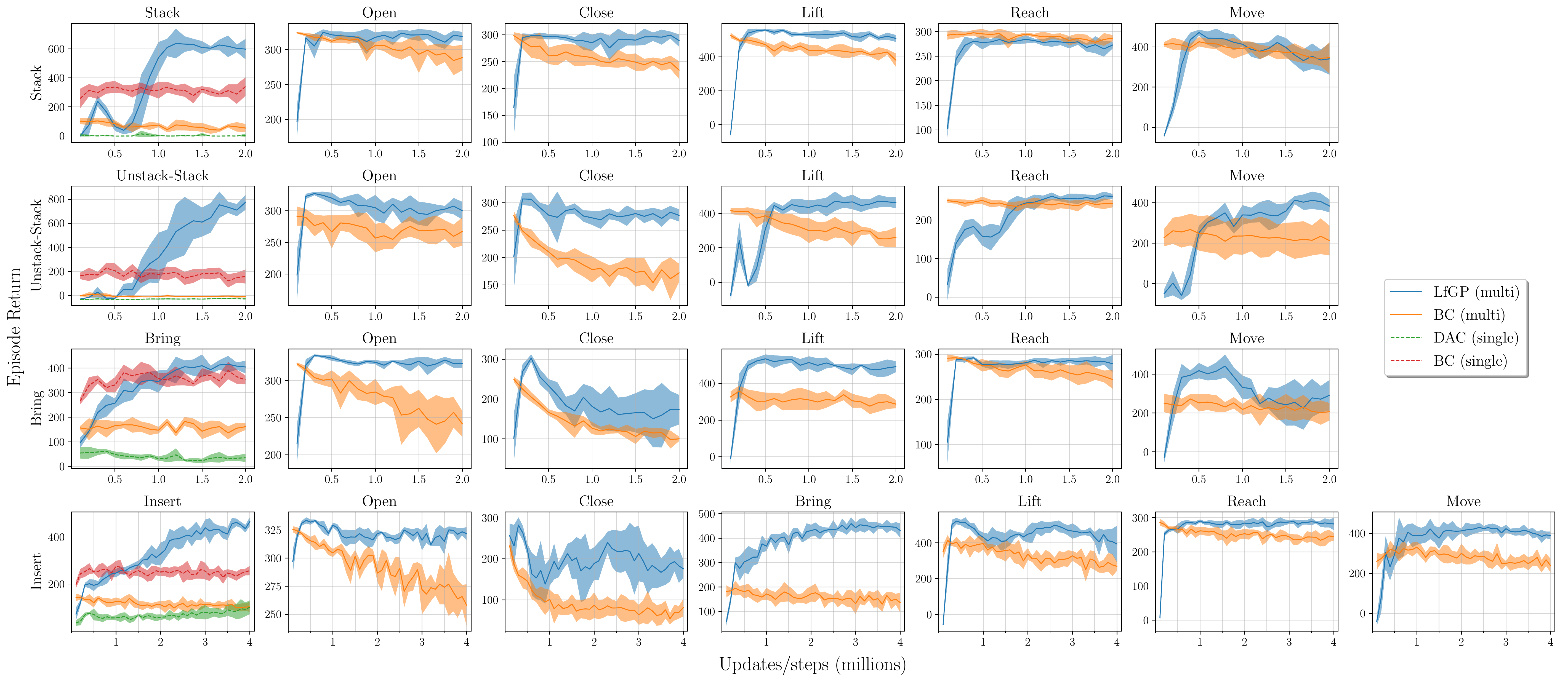}
	\caption{Episode return for LfGP compared with multitask baselines on all tasks. Shaded area corresponds to standard deviation.}
	\label{fig:multitask_return}
\end{figure*}

As previously stated, we generated hand-crafted reward functions for each of our tasks for the purpose of training our SAC-X experts.
Given that we have these rewards, we can also generate return plots corresponding to our results to add extra insight (see \cref{fig:main_return} and \cref{fig:multitask_return}).
The patterns displayed in these plots are, for the most part, quite similar to the success rate plots.
One notable exception is that there is an eventual increase in performance when training DAC on Insert, indicating that, perhaps for certain tasks, DAC alone can eventually make progress.
Nevertheless, it is clear that LfGP improves learning efficiency, and it is unclear whether DAC would plateau even if it was trained for a longer period.

\section{Model Architectures and Hyperparameters} \label{sec:hyperparameters}

All the single-task models share the same network architectures and all the multitask models share the same network architectures. 
All layers are initialized using the PyTorch default methods \citeapp{paszkePyTorchImperativeStyle2019}.

For the single-task variant, the policy is a fully-connected network with two hidden layers followed by ReLU activation. 
Each hidden layer consists of 256 hidden units. 
The output of the policy for LfGP and DAC is split into two vectors, mean $\hat{\mu}$ and variance $\hat{\sigma}^2$.
For both variants of BC, only the mean $\hat{\mu}$ output is used.
The vectors define a Gaussian distribution (i.e. $N(\hat{\mu}, \hat{\sigma}^2 \mathbf{I})$, where $\mathbf{I}$ is the identity matrix). When computing actions, we squash the samples using the tanh function and bound the actions to be in range $[-1, 1]$, as done in SAC \citeapp{haarnojaSoftActorCriticAlgorithms2019}.
The variance $\hat{\sigma}^2$ is computed by applying a softplus function followed by a sum with an epsilon $\epsilon =$ 1e-7 to prevent underflow: $\hat{\sigma}_i = \text{softplus}(\hat{x}_i) + \epsilon$.
The Q-functions are fully-connected networks with two hidden layers followed by ReLU activations.
Each hidden layer consists of 256 units.
The output of the Q-function is a scalar corresponding to the value estimate given the current state-action pair.
Finally, the discriminator is a fully-connected network with two hidden layers followed by tanh activations.
Each hidden layer consists of 256 units.
The output of the discriminator is a scalar logit to be used as an input to the sigmoid function.
The sigmoid function output can be viewed as the probability of the current state-action pair coming from the expert distribution.

For multitask variant, the policies and the Q-functions share their initial layers.
There are two shared, fully-connected layers followed by ReLU activations.
Each layer consists of 256 units.
The output of the last shared layer is then fed into the policies and Q-functions.
Each policy head and Q-function head corresponds to one task and has the same architecture: a two-layered fully-connected network followed by ReLU activations.
The output of the policy head corresponds to the parameters of a Gaussian distribution, as described previously.
Similarly, the output of the Q-function head corresponds to the value estimate.
Finally, the discriminator is a fully-connected network with two hidden layers followed by tanh activations.
Each hidden layer consists of 256 units.
The output of the discriminator is a vector, where the $i^\text{th}$ entry corresponds to the logit input to the sigmoid function for task $\tasks_i$.
The $i^\text{th}$ sigmoid function output corresponds to the probability of the current state-action pair coming from the expert distribution in task $\tasks_i$.

The hyperparameters for our experiments are listed in \cref{tab:hyperparameters_ail} and \cref{tab:hyperparameters_bc}.
In the early-stopping variant of BC, \textit{overfit tolerance} refers to the number of full dataset training epochs without an improvement in validation error before we stop training.
All models are optimized using Adam Optimizer \citeapp{kingmaAdamMethodStochastic2015} with PyTorch default values, unless specified otherwise.

\begin{table}[ht]
	\centering
	\small
	\caption{Hyperparameters for AIL algorithms across all tasks. 
        Parameters that do not appear in the original version of DAC are shown in \textcolor{blue}{blue}.}
	\begin{tabularx}{\columnwidth}{lcc}
        \toprule
            Algorithm & LfGP & DAC\\
        \midrule
        \midrule
            Total Interactions & \multicolumn{2}{c}{2M (4M for Insert)} \\
            Buffer Size & \multicolumn{2}{c}{2M (4M for Insert)} \\
		Buffer Warmup & \multicolumn{2}{c}{25k} \\
		Initial Exploration & \multicolumn{2}{c}{50k} \\
            Evaluations per task & \multicolumn{2}{c}{50} \\
            Evaluation frequency & \multicolumn{2}{c}{100k interactions} \\
        \midrule
        \midrule
		\textit{Intention} & &  \\
		$\gamma$ &  \multicolumn{2}{c}{0.99}  \\
		Batch Size & \multicolumn{2}{c}{256}  \\
		$Q$ Update Freq. & \multicolumn{2}{c}{1}  \\
		Target $Q$ Update Freq. & \multicolumn{2}{c}{1} \\
		$\pi$ Update Freq. & \multicolumn{2}{c}{1} \\
		Polyak Averaging & \multicolumn{2}{c}{1e-4} \\
            $Q$ Learning Rate & \multicolumn{2}{c}{3e-4} \\
		$\pi$ Learning Rate & \multicolumn{2}{c}{1e-5} \\
            $\alpha$ Learning Rate & \multicolumn{2}{c}{3e-4} \\
            Initial $\alpha$ & \multicolumn{2}{c}{1e-2} \\
            Target Entropy & \multicolumn{2}{c}{$-\text{dim}(a)=-4$} \\
            Max. Gradient Norm & \multicolumn{2}{c}{10} \\
            \textcolor{blue}{$\pi$ Weight Decay} & \multicolumn{2}{c}{\textcolor{blue}{1e-2}} \\
            \textcolor{blue}{$Q$ Weight Decay} & \multicolumn{2}{c}{\textcolor{blue}{1e-2}} \\
            \textcolor{blue}{$\buffer^E$ sampling proportion} & \multicolumn{2}{c}{\textcolor{blue}{0.1}} \\
            \textcolor{blue}{$\buffer^E$ sampling decay} & \multicolumn{2}{c}{\textcolor{blue}{0.99999}} \\
        \midrule
        \midrule
		\textit{Discriminator} & &  \\
		Learning Rate & \multicolumn{2}{c}{3e-4} \\
            Batch Size & \multicolumn{2}{c}{256} \\
            Gradient Penalty $\lambda$ & \multicolumn{2}{c}{10} \\
            \textcolor{blue}{Weight Decay} & \multicolumn{2}{c}{\textcolor{blue}{1e-2}} \\
            \textcolor{blue}{$(s_T, \boldsymbol{0})$ sampling bias} & \multicolumn{2}{c}{\textcolor{blue}{0.95}} \\
        \bottomrule
	\end{tabularx}
	\label{tab:hyperparameters_ail}
\end{table}

\begin{table}[ht]
	\centering
	\small
	\caption{Hyperparameters for LfGP schedulers.}
	\begin{tabularx}{\columnwidth}{lccc}
	    \toprule
            Scheduler & Learned & WRS & WRS + HC\\
            \midrule
            \midrule
            $\xi$ & 45 & N/A & N/A \\
		$\phi$ & 0.6 & N/A & N/A \\
		Initial Temp. & 360 & N/A & N/A \\
		Temp. Decay & 0.9995 & N/A & N/A \\
		Min. Temp. & 0.1 & N/A & N/A \\
            Main Task Rate & N/A & 0.5 & 0.5 \\
            Handcraft Rate & N/A & N/A & 0.5 \\
            \bottomrule
	\end{tabularx}
	\label{tab:hyperparameters_schedulers}
\end{table}

\begin{table}[ht]
	\centering
	\small
	\caption{Hyperparameters for BC algorithms (both single-task and multitask) across all tasks.}
	\begin{tabularx}{\columnwidth}{lcc}
	    \toprule
        Version & Main Results & Early Stopping  \\
        \midrule
        \midrule
		Batch Size & \multicolumn{2}{c}{256} \\
            Learning Rate & \multicolumn{2}{c}{1e-5} \\
            Weight Decay & \multicolumn{2}{c}{1e-2} \\
            Total Updates & 2M (4M for Insert) & N/A  \\
            Overfit Tolerance & N/A & 100 \\
        \bottomrule
	\end{tabularx}
	\label{tab:hyperparameters_bc}
\end{table}

\section{Open-Action and Close-Action\\ Distribution Matching}

There was one exception to the method we used for collecting our expert data.
Specifically, our Open-Gripper and Close-Gripper tasks required additional considerations.
It is worth reminding the reader that our Open-Gripper and Close-Gripper tasks were meant to simply open or close the gripper, respectively, while remaining reasonably close to either block.
If we were to use the approach described above verbatim, the Open-Gripper and Close-Gripper data would contain no $(s,a)$ pairs where the gripper actually released or grasped the block, instead immediately opening or closing the gripper while simply hovering near the blocks.
Perhaps unsurprisingly, this was detrimental to our algorithm's performance: as one example, an agent attempting to learn Stack would, if Open-Gripper was selected while the blue block was held above the green block, move the grasped blue block \textit{away} from the green block before dropping it on the tray.
This behaviour, of course, is not what we would want, but it better matches an expert distribution when the environment is reset in between each task execution.

To mitigate this, our Open-Gripper data actually contain a mix of each of the other sub-tasks called for the first 45 time steps, followed by a switch to Open-Gripper, ensuring that the expert dataset contains some degree of block-releasing, with the trade-off being that 50\% of the Open-Gripper expert data is specific to whatever the main task happens to be.
We left this additional detail out of our main paper for clarity, since it corresponds to only a small portion of the expert data (every other auxiliary task was fully reused).
Similarly, the Close-Gripper data calls Lift for 15 time steps before switching to Close-Gripper, ensuring that the Close-gripper dataset will contain a large proportion of data where the block is actually grasped.
For the Closer-gripper data, however, this modification did still allow data to be reused between main tasks.

\section{Attempted and Failed Experiments}

In this section, we provide a list of experiments and modifications that did not improve performance, in addition to the alternatives that did.

\begin{enumerate}
    \item \textbf{Pretraining with BC:} We attempted to pretrain LfGP using multitask BC, and then to transition to online learning with LfGP, but we found that this tended to produce significantly poorer final performance.
    Some existing work \citeapp{rajeswaran*LearningComplexDexterous2018, wuShapingRewardsReinforcement2020} has investigated transitioning from BC to online RL, but achieving this consistently, especially with off-policy RL, remains an open research problem.
    \item \textbf{Handcrafted Open-Gripper/Close-Gripper policies:} Given the simplicity of designing a reward function in these two cases, a natural question is whether Open-Gripper and Close-Gripper could use hand-crafted reward functions, or even hand-crafted policies, instead of these specialized datasets.
    In our experiments, both of these alternatives proved to be quite detrimental to our algorithm.
    \item \textbf{Penalizing Q values:} In our early experiments, we found that LfGP training progress was harmed by exploding Q values.
    This problem was particularly exacerbated when we added $\buffer^E$ sampling to our $Q$ and $\pi$ updates.
    It appears that this occurs because, at the beginning of training, the differences between discriminator outputs for expert data and non-expert data are so large that the bootstrap Q updates quickly jump to unrealistic values.
    We attempted to use various forms of Q penalties to resolve this, akin to Conservative Q Learning (CQL) \citeapp{kumarConservativeQLearningOffline2020}, but found that all of our modifications ultimately harmed final performance.
    Some of the things we tried, in addition to the CQL loss, were reducing $\gamma$ (.95, .9), clipping Q losses to -5, +5, smooth L1 loss, huber loss, increased gradient penalty $\lambda$ for $D$ (50, 100), decreased reward scaling (.1), more discriminator updates per $\pi$/$Q$ update (10), and weight decay in $D$ only (as is done in \citeapp{orsiniWhatMattersAdversarial2021}).
    We ultimately resolved exploding Q values by i) decreasing polyak averaging to a significantly lower value than is used in much other work (1e-4 as opposed to the SAC default of 5e-3), and ii) adding in weight decay (with a significantly higher value used than is used in other work) to $\pi$, $Q$, and $D$ training (which was required to not overfit with the reduced polyak averaging value).
    Without the added weight decay, performance started to plateau and eventually to decrease.
    \item \textbf{Higher Update-to-Data (UTD) Ratio:} Recent work in RL has started increasing the UTD ratio (i.e., increasing the number of policy/Q updates per environment interaction), with the goal of improving environment sample efficiency \citeapp{chenRandomizedEnsembledDouble2021}.
    We were actually able to increase this from 1 to 2 and achieve a marginal improvement in environment sample efficiency, but this also nearly doubled the running time of our experiments, so we opted not to include this modification in our final results.
    Higher values of the UTD ratio also caused our Q values to explode.   
\end{enumerate}

\section{Experimental Hardware}
For a list of the software we used in this work, see our code and instructions.
We used a number of different computers and GPUs when completing our experiments:
\begin{enumerate}
    \item GPU: NVidia Quadro RTX 8000, CPU: AMD - Ryzen 5950x 3.4 GHz 16-core 32-thread, RAM: 64GB, OS: Ubuntu 20.04.
    \item GPU: NVidia V100 SXM2, CPU: Intel Gold 6148 Skylake @ 2.4 GHz (only used 4 threads), RAM: 32GB, OS: CentOS 7.
    \item GPU: Nvidia GeForce RTX 2070, CPU: RYZEN Threadripper 2990WX, RAM: 32GB, OS: Ubuntu 20.04.
\end{enumerate}

%% file: root_combined.bbl
\begin{thebibliography}{10}
\providecommand{\url}[1]{#1}
\csname url@rmstyle\endcsname
\providecommand{\newblock}{\relax}
\providecommand{\bibinfo}[2]{#2}
\providecommand\BIBentrySTDinterwordspacing{\spaceskip=0pt\relax}
\providecommand\BIBentryALTinterwordstretchfactor{4}
\providecommand\BIBentryALTinterwordspacing{\spaceskip=\fontdimen2\font plus
\BIBentryALTinterwordstretchfactor\fontdimen3\font minus
  \fontdimen4\font\relax}
\providecommand\BIBforeignlanguage[2]{{%
\expandafter\ifx\csname l@#1\endcsname\relax
\typeout{** WARNING: IEEEtran.bst: No hyphenation pattern has been}%
\typeout{** loaded for the language `#1'. Using the pattern for}%
\typeout{** the default language instead.}%
\else
\language=\csname l@#1\endcsname
\fi
#2}}

\bibitem{rl_sandbox}
B.~Chan, ``{{RL}} sandbox,'' https://github.com/chanb/rl\_sandbox\_public,
  2020.

\bibitem{paszkePyTorchImperativeStyle2019}
A.~Paszke, \emph{et~al.}, ``{{PyTorch}}: {{An}} imperative style,
  high-performance deep learning library,'' in \emph{Advances in Neural Inf.
  Processing Systems 32}, H.~Wallach, H.~Larochelle, A.~Beygelzimer,
  F.~{dAlch{\'e}-Buc}, E.~Fox, and R.~Garnett, Eds.\hskip 1em plus 0.5em minus
  0.4em\relax {Curran Associates, Inc.}, 2019, pp. 8024--8035.

\bibitem{gulrajaniImprovedTrainingWasserstein2017}
I.~Gulrajani, F.~Ahmed, M.~Arjovsky, V.~Dumoulin, and A.~C. Courville,
  ``Improved {{Training}} of {{Wasserstein GANs}},'' in \emph{Conf. {{Neural
  Inf. Processing Systems}}}, I.~Guyon, \emph{et~al.}, Eds.\hskip 1em plus
  0.5em minus 0.4em\relax {Long Beach, USA}: {Curran Associates, Inc.}, Dec.
  2017, pp. 5767--5777.

\bibitem{kostrikovDiscriminatorActorCriticAddressingSample2019}
I.~Kostrikov, K.~K. Agrawal, D.~Dwibedi, S.~Levine, and J.~Tompson,
  ``Discriminator-{{Actor-Critic}}: {{Addressing Sample Inefficiency}} and
  {{Reward Bias}} in {{Adversarial Imitation Learning}},'' in \emph{Proc.
  {{Int. Conf.}} {{Learning Representations}} ({{ICLR}}'19)}, {New Orleans,
  USA}, May 2019.

\bibitem{kingmaAutoEncodingVariationalBayes2013}
D.~P. Kingma and M.~Welling, ``Auto-{{Encoding Variational Bayes}},''
  \emph{arXiv:1312.6114 [cs, stat]}, Dec. 2013.

\bibitem{fujimotoAddressingFunctionApproximation2018}
S.~Fujimoto, H.~{van Hoof}, and D.~Meger, ``Addressing {{Function Approximation
  Error}} in {{Actor-Critic Methods}},'' in \emph{Proc. 35th {{Int. Conf.}}
  {{Machine Learning}} ({{ICML}}'18)}, {Stockholm, Sweden}, Jul. 10--15 2018,
  pp. 1582--1591.

\bibitem{vanhasseltDeepReinforcementLearning2016}
H.~{van Hasselt}, A.~Guez, and D.~Silver, ``Deep {{Reinforcement Learning}}
  with {{Double Q-learning}},'' in \emph{{{AAAI Conf.}} {{Artificial
  Intelligence}}}, {Pheonix, USA}, Feb. 2016.

\bibitem{mnihHumanlevelControlDeep2015}
V.~Mnih, \emph{et~al.}, ``Human-level control through deep reinforcement
  learning,'' \emph{Nature}, vol. 518, no. 7540, pp. 529--533, Feb. 2015.

\bibitem{haarnojaSoftActorCriticAlgorithms2019}
T.~Haarnoja, \emph{et~al.}, ``Soft {{Actor-Critic Algorithms}} and
  {{Applications}},'' \emph{arXiv:1812.05905 [cs, stat]}, Jan. 2019.

\bibitem{kostrikovPyTorchImplementationsReinforcement2018}
I.~Kostrikov, ``{{PyTorch Implementations}} of {{Reinforcement Learning
  Algorithms}},'' https://github.com/ikostrikov/pytorch-a2c-ppo-acktr-gail,
  2018.

\bibitem{riedmillerLearningPlayingSolving2018}
M.~Riedmiller, \emph{et~al.}, ``Learning by {{Playing Solving Sparse Reward
  Tasks}} from {{Scratch}},'' in \emph{Proc. 35th {{Int. Conf.}} {{Machine
  Learning}} ({{ICML}}'18)}, {Stockholm, Sweden}, July 2018, pp. 4344--4353.

\bibitem{coumans2019}
E.~Coumans and Y.~Bai, ``{{PyBullet}}, a {{Python}} module for physics
  simulation for games, robotics and machine learning,'' http://pybullet.org,
  2016.

\bibitem{kingmaAdamMethodStochastic2015}
D.~P. Kingma and J.~Ba, ``Adam: {{A Method}} for {{Stochastic Optimization}},''
  in \emph{Proc. Int. Conf. Learning Representations ({{ICLR}}'15)}, {San
  Diego, USA}, May 7--9 2015.

\bibitem{rajeswaran*LearningComplexDexterous2018}
A.~Rajeswaran*, \emph{et~al.}, ``Learning {{Complex Dexterous Manipulation}}
  with {{Deep Reinforcement Learning}} and {{Demonstrations}},'' in \emph{Proc.
  {{Robotics}}: {{Science}} and {{Systems}} ({{RSS}}'18)}, {Pittsburgh, USA},
  Jun. 26--30 2018.

\bibitem{wuShapingRewardsReinforcement2020}
Y.~Wu, M.~Mozifian, and F.~Shkurti, ``Shaping {{Rewards}} for {{Reinforcement
  Learning}} with {{Imperfect Demonstrations}} using {{Generative Models}},''
  \emph{arXiv:2011.01298 [cs]}, Nov. 2020.

\bibitem{kumarConservativeQLearningOffline2020}
A.~Kumar, A.~Zhou, G.~Tucker, and S.~Levine, ``Conservative {{Q-Learning}} for
  {{Offline Reinforcement Learning}},'' \emph{arXiv:2006.04779 [cs, stat]},
  Aug. 2020.

\bibitem{orsiniWhatMattersAdversarial2021}
M.~Orsini, \emph{et~al.}, ``What {{Matters}} for {{Adversarial Imitation
  Learning}}?'' in \emph{Conf. {{Neural Inf. Processing Systems}}}, June 2021.

\bibitem{chenRandomizedEnsembledDouble2021}
X.~Chen, C.~Wang, Z.~Zhou, and K.~Ross, ``Randomized {{Ensembled Double
  Q-Learning}}: {{Learning Fast Without}} a {{Model}},'' \emph{arXiv:2101.05982
  [cs]}, Mar. 2021.

\end{thebibliography}


\begin{thebibliography}{10}
\providecommand{\url}[1]{#1}
\csname url@rmstyle\endcsname
\providecommand{\newblock}{\relax}
\providecommand{\bibinfo}[2]{#2}
\providecommand\BIBentrySTDinterwordspacing{\spaceskip=0pt\relax}
\providecommand\BIBentryALTinterwordstretchfactor{4}
\providecommand\BIBentryALTinterwordspacing{\spaceskip=\fontdimen2\font plus
\BIBentryALTinterwordstretchfactor\fontdimen3\font minus
  \fontdimen4\font\relax}
\providecommand\BIBforeignlanguage[2]{{%
\expandafter\ifx\csname l@#1\endcsname\relax
\typeout{** WARNING: IEEEtran.bst: No hyphenation pattern has been}%
\typeout{** loaded for the language `#1'. Using the pattern for}%
\typeout{** the default language instead.}%
\else
\language=\csname l@#1\endcsname
\fi
#2}}

\bibitem{suttonReinforcementLearningIntroduction2018}
R.~S. Sutton and A.~G. Barto, \emph{Reinforcement Learning: {{An}}
  Introduction}, 2nd~ed.\hskip 1em plus 0.5em minus 0.4em\relax {MIT press},
  2018.

\bibitem{bellemareUnifyingCountBasedExploration2016}
M.~Bellemare, S.~Srinivasan, G.~Ostrovski, T.~Schaul, D.~Saxton, and R.~Munos,
  ``Unifying {{Count-Based Exploration}} and {{Intrinsic Motivation}},'' in
  \emph{Conf. {{Neural Inf. Processing Systems}}}, vol.~29, Dec. 2016.

\bibitem{nairOvercomingExplorationReinforcement2018}
A.~Nair, B.~McGrew, M.~Andrychowicz, W.~Zaremba, and P.~Abbeel, ``Overcoming
  {{Exploration}} in {{Reinforcement Learning}} with {{Demonstrations}},'' in
  \emph{Proc. 2018 {{IEEE Int. Conf.}} {{Robotics}} and {{Automation}}
  ({{ICRA}}'18)}, {Brisbane, Australia}, May 2018, pp. 6292--6299.

\bibitem{ngShapingPolicySearch2003}
A.~Y. Ng and M.~I. Jordan, ``Shaping and policy search in reinforcement
  learning,'' Ph.D. dissertation, University of California, Berkeley, 2003.

\bibitem{ngAlgorithmsInverseReinforcement2000}
A.~Ng and S.~Russell, ``Algorithms for inverse reinforcement learning,'' in
  \emph{Int. {{Conf.}} {{Machine Learning}} ({{ICML}}'00)}, July 2000, pp.
  663--670.

\bibitem{hoGenerativeAdversarialImitation2016}
J.~Ho and S.~Ermon, ``Generative {{Adversarial Imitation Learning}},'' in
  \emph{Conf. {{Neural Inf. Processing Systems}}}, {Barcelona, Spain}, Dec.
  5--11 2016, pp. 4565--4573.

\bibitem{kostrikovDiscriminatorActorCriticAddressingSample2019}
I.~Kostrikov, K.~K. Agrawal, D.~Dwibedi, S.~Levine, and J.~Tompson,
  ``Discriminator-{{Actor-Critic}}: {{Addressing Sample Inefficiency}} and
  {{Reward Bias}} in {{Adversarial Imitation Learning}},'' in \emph{Proc.
  {{Int. Conf.}} {{Learning Representations}} ({{ICLR}}'19)}, {New Orleans,
  USA}, May 2019.

\bibitem{fuLearningRobustRewards2018}
J.~Fu, K.~Luo, and S.~Levine, ``Learning {{Robust Rewards}} with
  {{Adverserial}} inverse {{Reinforcement Learning}},'' in \emph{Proc. {{Int.
  Conf.}} {{Learning Representations}} ({{ICLR}}'18)}, {Vancouver, Canada},
  Apr. 30--May 3 2018.

\bibitem{orsiniWhatMattersAdversarial2021}
M.~Orsini, \emph{et~al.}, ``What {{Matters}} for {{Adversarial Imitation
  Learning}}?'' in \emph{Conf. {{Neural Inf. Processing Systems}}}, June 2021.

\bibitem{ecoffetFirstReturnThen2021}
A.~Ecoffet, J.~Huizinga, J.~Lehman, K.~O. Stanley, and J.~Clune, ``First
  return, then explore,'' \emph{Nature}, vol. 590, no. 7847, pp. 580--586, Feb.
  2021.

\bibitem{ablettLearningGuidedPlay2021}
T.~Ablett, B.~Chan, and J.~Kelly, ``Learning from {{Guided Play}}: {{A
  Scheduled Hierarchical Approach}} for {{Improving Exploration}} in
  {{Adversarial Imitation Learning}},'' in \emph{Proc. {{Neural Inf. Processing
  Systems}} ({{NeurIPS}}'21) {{Deep Reinforcement Learning Workshop}}}, Dec.
  2021.

\bibitem{riedmillerLearningPlayingSolving2018}
M.~Riedmiller, \emph{et~al.}, ``Learning by {{Playing Solving Sparse Reward
  Tasks}} from {{Scratch}},'' in \emph{Proc. 35th {{Int. Conf.}} {{Machine
  Learning}} ({{ICML}}'18)}, {Stockholm, Sweden}, July 2018, pp. 4344--4353.

\bibitem{lynchLearningLatentPlans2019}
C.~Lynch, \emph{et~al.}, ``Learning {{Latent Plans}} from {{Play}},'' in
  \emph{Conf. {{Robot Learning}} ({{CoRL}}'19)}, 2019.

\bibitem{guptaRelayPolicyLearning2019}
A.~Gupta, V.~Kumar, C.~Lynch, S.~Levine, and K.~Hausman, ``Relay {{Policy
  Learning}}: {{Solving Long Horizon Tasks Via Imitation}} and {{Reinforcement
  Learning}},'' in \emph{Conf. {{Robot Learning}} ({{CoRL}}'19)}, 2019.

\bibitem{haarnojaSoftActorCriticOffPolicy2018}
T.~Haarnoja, A.~Zhou, P.~Abbeel, and S.~Levine, ``Soft {{Actor-Critic}}:
  {{Off-Policy Maximum Entropy Deep Reinforcement Learning}} with a
  {{Stochastic Actor}},'' in \emph{Proc. 35th Int. Conf. Machine Learning
  ({{ICML}}'18)}, {Stockholm, Sweden}, July 2018, pp. 1861--1870.

\bibitem{vecerikLeveragingDemonstrationsDeep2018}
M.~Vecerik, \emph{et~al.}, ``Leveraging {{Demonstrations}} for {{Deep
  Reinforcement Learning}} on {{Robotics Problems}} with {{Sparse Rewards}},''
  Oct. 2018.

\bibitem{kalashnikovQTOptScalableDeep2018}
D.~Kalashnikov, \emph{et~al.}, ``{{QT-Opt}}: {{Scalable Deep Reinforcement
  Learning}} for {{Vision-Based Robotic Manipulation}},''
  \emph{arXiv:1806.10293 [cs, stat]}, June 2018.

\bibitem{mandlekarWhatMattersLearning2021}
A.~Mandlekar, \emph{et~al.}, ``What {{Matters}} in {{Learning}} from {{Offline
  Human Demonstrations}} for {{Robot Manipulation}},'' in \emph{Conf. {{Robot
  Learning}}}, Nov. 2021.

\bibitem{hussenotHyperparameterSelectionImitation2021}
L.~Hussenot, \emph{et~al.}, ``Hyperparameter {{Selection}} for {{Imitation
  Learning}},'' in \emph{Proc. 38th {{Int. Conf.}} {{Machine Learning}}
  ({{ICML}}'21)}, July 2021, pp. 4511--4522.

\bibitem{fuVariationalInverseControl2018}
J.~Fu, A.~Singh, D.~Ghosh, L.~Yang, and S.~Levine, ``Variational {{Inverse
  Control}} with {{Events}}: {{A General Framework}} for {{Data-Driven Reward
  Definition}},'' in \emph{Conf. {{Neural Inf. Processing Systems}}},
  {Montreal, Canada}, Dec. 2018.

\bibitem{cabiIntentionalUnintentionalAgent2017}
S.~Cabi, \emph{et~al.}, ``The {{Intentional Unintentional Agent}}: {{Learning}}
  to {{Solve Many Continuous Control Tasks Simultaneously}},'' in \emph{Conf.
  {{Robot Learning}} ({{CoRL}}'17)}, {Mountain View, USA}, Nov. 2017.

\bibitem{zolnaTaskRelevantAdversarialImitation2021}
K.~Zolna, \emph{et~al.}, ``Task-{{Relevant Adversarial Imitation Learning}},''
  in \emph{Proc. 2020 {{Conf.}} {{Robot Learning}}}, Oct. 2021, pp. 247--263.

\bibitem{rossReductionImitationLearning2011}
S.~Ross, G.~J. Gordon, and D.~Bagnell, ``A {{Reduction}} of {{Imitation
  Learning}} and {{Structured Prediction}} to {{No-Regret Online Learning}},''
  in \emph{Proc. 14th {{Int. Conf.}} {{Artificial Intelligence}} and
  {{Statistics}} ({{AISTATS}}'11)}, {Fort Lauderdale, USA}, Apr. 2011, pp.
  627--635.

\bibitem{ablettSeeingAllAngles2021}
T.~Ablett, Y.~Zhai, and J.~Kelly, ``Seeing {{All}} the {{Angles}}: {{Learning
  Multiview Manipulation Policies}} for {{Contact-Rich Tasks}} from
  {{Demonstrations}},'' in \emph{Proc. {{IEEE}}/{{RSJ Int. Conf.}}
  {{Intelligent Robots}} and {{Systems}} ({{IROS}}'21)}, {Prague, Czech
  Republic}, Sep. 2021.

\bibitem{abbeelApprenticeshipLearningInverse2004}
P.~Abbeel and A.~Y. Ng, ``Apprenticeship learning via inverse reinforcement
  learning,'' in \emph{Int. {{Conf.}} {{Machine Learning}}
  ({{ICML}}'04)}.\hskip 1em plus 0.5em minus 0.4em\relax {Banff, Canada}: {ACM
  Press}, 2004.

\bibitem{ablettFightingFailuresFIRE2020}
T.~Ablett, F.~Mari{\'c}, and J.~Kelly, ``Fighting {{Failures}} with {{FIRE}}:
  {{Failure Identification}} to {{Reduce Expert Burden}} in
  {{Intervention-Based Learning}},'' \emph{arXiv:2007.00245 [cs]}, Aug. 2020.

\bibitem{hausmanMultiModalImitationLearning2017}
K.~Hausman, Y.~Chebotar, S.~Schaal, G.~Sukhatme, and J.~Lim, ``Multi-{{Modal
  Imitation Learning}} from {{Unstructured Demonstrations}} using {{Generative
  Adversarial Nets}},'' in \emph{Conf. {{Neural Inf. Processing Systems}}}, May
  2017.

\bibitem{suttonMdpsSemiMdpsFramework1999}
R.~S. Sutton, D.~Precup, and S.~Singh, ``Between {{MDPs}} and {{Semi-MDPs}}:
  {{A Framework}} for {{Temporal Abstraction}} in {{Reinforcement Learning}},''
  \emph{Artificial Intelligence}, vol. 112, no. 1-2, pp. 181--211, Aug. 1999.

\bibitem{nachumWhyDoesHierarchy2019}
O.~Nachum, H.~Tang, X.~Lu, S.~Gu, H.~Lee, and S.~Levine, ``Why {{Does
  Hierarchy}} ({{Sometimes}}) {{Work So Well}} in {{Reinforcement Learning}}?''
  in \emph{Proc. {{Neural Inf. Processing Systems}} ({{NeurIPS}}'19) {{Deep
  Reinforcement Learning Workshop}}}, Sep. 2019.

\bibitem{hendersonOptionGANLearningJoint2018}
P.~Henderson, W.-D. Chang, P.-L. Bacon, D.~Meger, J.~Pineau, and D.~Precup,
  ``{{OptionGAN}}: {{Learning Joint Reward-Policy Options Using Generative
  Adversarial Inverse Reinforcement Learning}},'' in \emph{Proc. {{AAAI Conf.}}
  {{Artificial Intelligence}} ({{AAAI}}'18)}, no.~1, Apr. 2018.

\bibitem{sharmaDirectedInfoGAILLearning2019}
M.~Sharma, A.~Sharma, N.~Rhinehart, and K.~M. Kitani, ``Directed-{{Info GAIL}}:
  {{Learning Hierarchical Policies}} from {{Unsegmented Demonstrations}} using
  {{Directed Information}},'' in \emph{Int. {{Conf.}} {{Learning
  Representations}} ({{ICLR}}'19)}, May 2019.

\bibitem{jingAdversarialOptionAwareHierarchical2021}
M.~Jing, \emph{et~al.}, ``Adversarial {{Option-Aware Hierarchical Imitation
  Learning}},'' in \emph{Proc. 38th {{Int. Conf.}} {{Machine Learning}}
  ({{ICML}}'21)}, July 2021, pp. 5097--5106.

\bibitem{codevillaEndtoEndDrivingConditional2018}
F.~Codevilla, M.~M{\"u}ller, A.~L{\'o}pez, V.~Koltun, and A.~Dosovitskiy,
  ``End-to-{{End Driving Via Conditional Imitation Learning}},'' in \emph{Proc.
  {{IEEE}} Int. Conf. Robotics and Automation ({{ICRA}}'18)}, {Brisbane,
  Australia}, May 21--25 2018, pp. 4693--4700.

\bibitem{pateriaHierarchicalReinforcementLearning2021}
S.~Pateria, B.~Subagdja, A.-h. Tan, and C.~Quek, ``Hierarchical {{Reinforcement
  Learning}}: {{A Comprehensive Survey}},'' \emph{ACM Computing Surveys},
  vol.~54, no.~5, pp. 109:1--109:35, June 2021.

\bibitem{guptaResetFreeReinforcementLearning2021}
A.~Gupta, \emph{et~al.}, ``Reset-{{Free Reinforcement Learning}} via
  {{Multi-Task Learning}}: {{Learning Dexterous Manipulation Behaviors}}
  without {{Human Intervention}},'' in \emph{Proc. 2021 {{IEEE Int. Conf.}}
  {{Robotics}} and {{Automation}} ({{ICRA}}'21)}, Apr. 2021.

\end{thebibliography}
